\documentclass[10pt,twocolumn,letterpaper]{article}

\usepackage[pagenumbers]{cvpr} % To force page numbers, e.g. for an arXiv version

\usepackage{graphicx}
\usepackage{amsmath}
\usepackage{amssymb}
\usepackage{booktabs}

\usepackage[pagebackref,breaklinks,colorlinks]{hyperref}

\usepackage[capitalize]{cleveref}
\crefname{section}{Sec.}{Secs.}
\Crefname{section}{Section}{Sections}
\Crefname{table}{Table}{Tables}
\crefname{table}{Tab.}{Tabs.}

\def\cvprPaperID{*****} % *** Enter the CVPR Paper ID here
\def\confName{CVPR}
\def\confYear{2022}

\begin{document}

\title{Estimating Neural Reflectance Field from Radiance Field using Tree Structures}

\author{
Xiu Li$^{1,2*}$\hspace{22pt}
Xiao Li$^1$\hspace{22pt}
Yan Lu$^1$\\
$^1$Microsoft Research Asia\hspace{22pt}
$^2$Tencent}

\maketitle

%%%%%%%%% ABSTRACT
\begin{abstract}
% Our method
We present a new method for estimating the Neural \textbf{Reflectance} Field (NReF) of an object from a set of posed multi-view images under unknown lighting. NReF represents 3D geometry and appearance of objects in a disentangled manner, and are hard to be estimated from images only.
Our method solve this problem by exploiting the Neural \textbf{Radiance} Field (NeRF) as a proxy representation, from which we perform further decomposition.
% Key observation
A high-quality NeRF decomposition relies on good geometry information extraction as well as good prior terms to properly resolve ambiguities between different components.
To extract high-quality geometry information from radiance fields, we re-design a new ray-casting based method for surface point extraction. To efficiently compute and apply prior terms, we convert different prior terms into different type of filter operations on the surface extracted from radiance field.
We then employ two type of auxiliary data structures, namely Gaussian KD-tree and octree, to support fast querying of surface points and efficient computation of surface filters during training.
%These priors and strategies essentially require non-local, geometric-aware information to compute, which cannot be extracted trivially from implicit field functions like NeRF and NReF.
% Our technical solution
%To address this, we maintain a memory bank module that stores a discrete version of the NReF with high-efficient data structures.
%By introducing the memory bank as an auxiliary function, different priors can be efficiently computed and dynamic ray-sampling strategies can be easily applied. 
Based on this, we design a multi-stage decomposition optimization pipeline for estimating neural reflectance field from neural radiance fields.
% Results
Extensive experiments show our method outperforms other state-of-the-art methods on different data, and enable high-quality free-view relighting as well as material editing tasks.
\end{abstract}

\let\thefootnote\relax\footnotetext{* This work was done during Xiu Li's internship at Microsoft Research Asia.}
%%%%%%%%% BODY TEXT
\section{Introduction}
\begin{figure*}[!ht]
     \centering
     \includegraphics[width=0.8\textwidth]{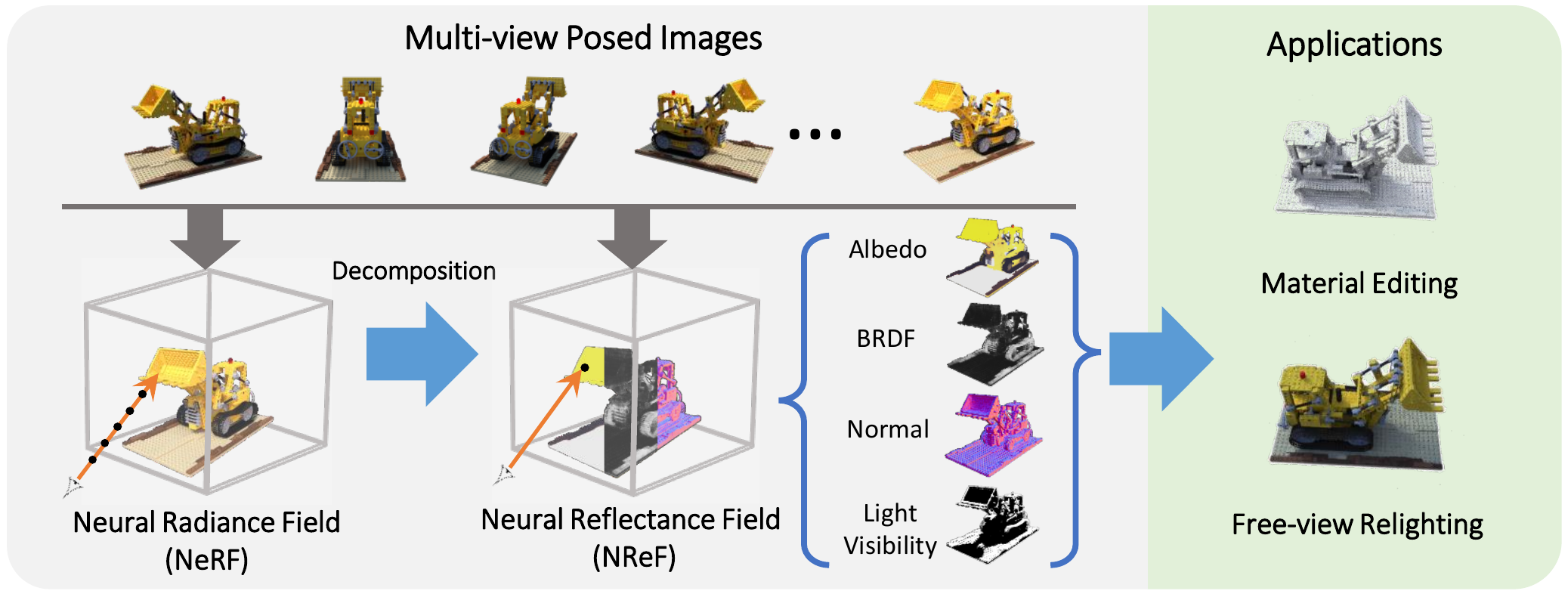}
     \caption{Given a set of multi-view posed images of an object with unknown illumination only (left-top), we estimate a neural reflectance field (NReF, mid-bottom) which decompose the neural radiance field (NeRF, left-bottom) of the object into fields of albedo, BRDF, surface normals and light visibility. NReF enables photo-realistic 3D editing tasks such as material editing and object relighting (right).}
     \label{fig:idea_overview}
\end{figure*}
\label{sec:intro}
% Introduce the problem
The problem of digitally reproducing, editing, and photo-realistically synthesizing an object's 3D shape and appearance is a fundamental research topic with many applications, ranging from virtual conferencing to augment reality.
Despite its usefulness, this topic is very challenging because of its inherently highly ill-posed nature and a highly non-linear optimization process, due to the complex interplay of shape, reflectance, and lighting\cite{Xia:2016:RSS} in the observations.
Typical inverse-rendering approaches\cite{kang2019learning,bi2020deep,gao2020deferred} rely on either dedicated capture devices, active lighting, or restrictive assumptions on target geometry and/or materials.

Recently, the pioneering work of NeRF~\cite{mildenhall2020nerf} has shown great advances in 3D reconstruction from a set of posed multi-view images without additional setups.
NeRF represents a \textbf{radiance} field for a given object using neural network as an implicit function.
A radiance field is suitable for view synthesis but cannot support further  manipulation tasks due to its entanglement of reflectance and lighting.
To fully solve the inverse rendering problem and supports manipulation, a more suitable representation is \textbf{reflectance} field\cite{bi2020neural}, \cite{bi2020deep}, which represents shape, reflectance and lighting in a disentangled manner.

Given the surprisingly high reconstruction quality and simple cature setup of neural radiance fields (i.e., NeRF), a few recent works (\cite{bi2020neural,nerv2021,boss2021nerd,zhang2021nerfactor}) have been attempted extending neural representations to reflectance fields.
%a naturally idea is to extend it to model reflectance fields. Indeed, a few recent works (\cite{bi2020neural}, \cite{nerv2021}, \cite{boss2021nerd}, \cite{zhang2021nerfactor}) have been proposed methods for neural reflectance field estimation and showing promising results. 
Yet, some of those methods still need additional inputs such lighting information; other methods without additional input requirements, are still struggling at fully resolving the high-complexity of inverse-rendering optimization, producing noticeable artifacts and/or degenerated results.
Thus, a set of questions naturally come up:
\textbf{Can we really achieve high quality estimation of reflectance fields with neural representations? And if possible, what is the key for an effective and robust estimation of neural reflectance fields using only posed multi-view images with unknown lighting?}

In this paper, we provide positive answers to the two questions, by proposing a new method that estimates a neural \textbf{reflectance} field for a given object from only a set of multi-view images under unknown lighting.
Inspired by\cite{zhang2021nerfactor}, we formulate this problem as a two-stage optimization: An initial NeRF training stage and a NeRF decomposition stage.
The pre-trained NeRF gives a plausible initialization for object shape but reflectance properties and lighting are still entangled.
We then train a set of neural networks to represent implicit fields of reflectance, surface normal, and lighting visibility respectively. Fig.~\ref{fig:idea_overview} demonstrates our idea.
To avoid confusion, we will use NReF for {\bf N}eural {\bf Re}flectance {\bf F}ield afterwards.

% Key challenges
A key challenge of decomposing neural radiance field into neural reflectance fields is to correctly extract geometry information as priors from the pre-trained NeRF. Unlike radiance fields that generate the final renderings with volumetric integration, a reflectance field only computes its rendering results on surface points of the corresponded object. Thus, a robust and accurate surface point extraction method is required for computing shading color and geometry visibility terms.
However, current surface point extraction method based on volumetric density integration, using by most NeRF-based methods (\cite{mildenhall2020nerf}, ~\cite{zhang2021nerfactor}), often produces too surface extraction results for a robust geometry initialization, as we will shown later.
To alleviate this problem, we revisit their method and amend surface the point extraction process by proposing an effective strategy based on ray-casting. To support fast point querying during training, we construct an octree on densely sampled point cloud from NeRF.

The second key challenge of high-quality neural reflectance field optimization under unknown lighting is to resolve ambiguities due to its intrinsically ill-posed nature.
Previous image-based reflectance decomposition methods (\cite{BarronTPAMI2015}) have shown that adding suitable smoothness and parsimony prior terms is crucial to resolve shading/albedo ambiguity.
Our key observation is that, adding different type of priors mentioned above during training, can be unified as applying different type of filters on the geometry surface. However, applying such filters are non-trivial for neural reflectace field as the surface are only defined with implicit functions. To address this issue, we exploit the idea of Gaussian KD-tree (\cite{adams2009gaussian}) to efficiently compute a discrete sampled approximation of all prior terms, and employ a commitment loss to propagate the prior back into the implicit fields. In this way, we are able to add suitable priors for decomposing reflectance and shading and significantly improving the quality of neural reflectance field estimation.

Based on the two auxiliary tree-based data structures, we design an optimization pipeline with carefully considerations on surface extraction, prior terms, and importance sampling of lighting. 
Our pipeline enables the estimation of high-quality neural reflectance fields with only multi-view posed images under unknown lighting as input.
We validate and demonstrate the strength of our method with extensive experiments on both synthetic and real data. We also apply our method to manipulation tasks such as relighting and material editing.

To summarize, our contributions are as follows:
\begin{itemize}
    \item A novel approach for estimating reflectance field of 3D objects using only multi-view posed images under uncontrolled, unknown lighting.
    \item A new method to extract surfaces point from pre-trained radiance fields with reduced noise.
    \item A dedicate designed optimization pipeline that decomposes a neural radiance fields into neural reflectance fields to support manipulation tasks.
\end{itemize}

\section{Related Works}
\label{sec:related_works}
\noindent{\textbf{Inverse Rendering.}}
The task of inverse rendering is to decompose an observed image of a given object into geometry, appearance properties and lighting conditions, such that the components follow the physical imaging process.
Since the decomposition is intrinsically an ill-posed problem, most prior approaches address this problem by adding strong assumptions on object shape (\cite{BarronTPAMI2015,li2017modeling,ye2018single,gao2019deep,li2018materials,deschaintre2019flexible}), exploiting additional information of shape or lighting (\cite{dong2014appearance,nerv2021,bi2020deep}), or designing dedicated devices for controllable capturing (\cite{kang2019learning,ma2021free}).
Our method only uses multi-view images as input and has less restriction on shapes/materials.\\

\noindent{\textbf{Neural 3D Representations.}}
Recently, the neural representation of 3D scenes has attracted considerable attention in the literature(\cite{mildenhall2020nerf,barron2021mip,sitzmann2020implicit,park2019deepsdf,deng2021deformed}). These methods exploit multi-layer perceptrons to represent implicit fields such as sign distance functions for surface or volumetric radiance fields, known as Neural Fields.
%thus inherently encode 3D information in a view-consistent manner.
% NeRF and its variants
Our method builds upon the neural radiance field (NeRF) for 3D representation. NeRF~\cite{mildenhall2020nerf} and its variants have surpassed previous state-of-the-art methods on novel view synthesis tasks; however, NeRF cannot support various editing tasks because it models radiance fields as a ``black-box".
Our work takes one step further towards opening this "black-box" by providing a method to decompose NeRF into shape, reflectance and lighting, enabling editing tasks.
Some prior arts also attempt to model reflectance fields with neural networks.
NeRV~\cite{nerv2021} proposed a method that estimates reflectance fields from multi-view images with known lighting. Bi et el.~\cite{bi2020neural} estimate reflectance fields from images captured with a collocated camera-light setup. Our method does not require lighting conditions as prior information.
NeRD~\cite{boss2021nerd} and PhySG~\cite{zhang2021physg} directly solve reflectance fields from multi-view posed images with unknown illumination.
Both NeRD and PhySG do not take light visibility into account and are unable to simulate any lighting occlusion or shadowing effects.
We address this issue by modeling the light visibility field in our decomposition.
The most similar work to us is NeRFactor~\cite{zhang2021nerfactor} which also decomposes a reflectance field from a pre-trained NeRF.
A key drawback of NeRFactor is their limited quality. Overall, NeRFactor tends to output over smoothed normal, less disentangled albedo/shading, and degenerated specular components.
Our method greatly improve the quality of neural reflectance field by improving the surface point extraction, correctly handling dynamic importance sampling, and adding additional priors. These improvements cannot be trivially implemented without our introducing of tree-based data structures and carefully designed training strategies.\\
%[n]NeROIC: Neural Object Capture and Rendering from Online Image Collections

\noindent{\textbf{Data structures for neural representations.}} The octree data structure have been used in several works to accelerate training and/or rendering of neural radiance fields (\cite{liu2020neural},~\cite{yu2021plenoctrees}). The method of Gaussian KD-Tree~\cite{adams2009gaussian} has been used for accelerating a broad class of non-linear filters that includes the bilateral, non-local means, and other related filters. Both data structures plays an important role in our method during NReF training: the octree gives us the ability to query extracted surface points on the fly for computing geometric visibility terms, and the Gaussian KD-tree enables us to apply different prior term in a unified way by filtering high-dimensional features on object surfaces.
%\noindent\textbf{Local/global Optimization}
%The method of local/global optimization, also known as block coordinate descent or alternating optimization, has been successfully employed in some computer graphic problems, such as as-rigid-as-possible surface mesh deformation~\cite{10.2312:SGP:SGP07:109-116}, fast simulation of mass-spring systems~\cite{Liu:2013:FSM}, etc.
%We extend this methodology to NeRF literature by introducing memory bank as auxiliary function to support high-quality NeRF decomposition.
%These methods surpassed previous state- of-the-art methods in novel view interpolation and achieved photo-realistic results in most cases
%Neural Volume, NeRF;
%NeRV, NeRD, Neural Reflectance fields;
%NeRFactor

%\noindent{\textbf{Memory bank in training neural
%networks}}
%MoCo; BYOL

\section{Method}
\label{sec:method}
Our goal is to estimate a neural reflectance field (NReF), given only $n$ multi-view posed images $\{I_k | k=1...n\}$ with unknown lighting as observations.
A NReF $f(\mathbf{x})$ represents the shape, light, and reflectance properties of an object at any 3D location $\mathbf{x}$ on its opaque surface. We parameterize NReF with a set of multi-layer perceptron (MLP) networks and solve the NReF estimation with a `NeRF decomposition' approach. A NeRF MLP is first is trained with the same set of inputs (section~\ref{sec:method:nerf_to_nref}) and the initial surface geometry is extracted from it with a novel ray-casting based approach, accelerated with octree (section~\ref{sec:method:surface_extraction}).
The decomposition itself relies on a set of priors to resolve ambiguities that are non-trivial to employ with neural implicit field representations only. We address this issue with a Gaussian KD-tree that converts priors into surface filtering operations (section~\ref{sec:method:priors}).
Finally, we introduce our multi-stage NReF decomposition pipeline with implementation details (section~\ref{sec:method:nref_decomposition}).

\subsection{From Radiance to Reflectance}
\label{sec:method:nerf_to_nref}
We begin by training a Neural Radiance Field (NeRF) following the same procedure in~\cite{mildenhall2020nerf}.
In NeRF, the rendered color $C(\mathbf{r})$ of the camera ray $\mathbf{r}(t)=\mathbf{o}+t\mathbf{d}$ is generated by querying and blending the radiance $L_o(\boldsymbol\omega_v,\mathbf{r}(t))$ according to the volume density value $\sigma(\mathbf{r}(t))$ alongside $\mathbf{r}$(t) via
\begin{equation}
    C(\mathbf{r}) = \int_0^{\infty} \frac{\partial T(\mathbf{r}(t))}{\partial t} L_o(\boldsymbol\omega_v,\mathbf{r}(t))dt
    \label{equ:nerf_render}
\end{equation}
where
\begin{equation}
T(\mathbf{r}(t)) = 1-\exp\left(-\int_0^t \sigma(\mathbf{r}(t))ds \right)
\label{equ:transmit}
\end{equation}
Here, $\boldsymbol\omega_v = -\mathbf{d}/\lVert\mathbf{d}\rVert$ is the normalized view direction, and $T(\mathbf{r}(t))$ is the transmittance function. 
NeRF works well for view synthesis since it already learned reasonable shape via the volume density $\sigma(t)$; however, it is not suitable for other manipulations of shading effects because reflectance and lighting are still entangled. To enable control over those factors, we formulate a decomposition problem for estimating NReF as follows.\\

\noindent{\textbf{Reflectance field formulation}}~
The relationship of radiance, shape, reflectance, and lighting at surface point $\mathbf{x}$ from direction $\boldsymbol\omega_v$ is given by the rendering equation~(\cite{kajiya1986rendering}):
\begin{equation}
L_o(\boldsymbol\omega_v,\mathbf{x}) = 
\int f_r(\boldsymbol\omega_v,\boldsymbol\omega_i,\mathbf{x})L_i(\boldsymbol\omega_i,\mathbf{x})\max(f_n(\mathbf{x})\cdot\boldsymbol\omega_i,0) d\boldsymbol\omega_i
\label{equ:rendering}
\end{equation}
where $f_r(\cdot)$ is the Bidirectional Reflectance Distribution Function (BRDF), $L_i(\boldsymbol\omega_i,\mathbf{x})$ is the incident light at direction $\boldsymbol\omega_i$ and $f_n(\cdot)$ is the surface normal. We further assume light sources are far-field and decompose the lighting $L_i(\cdot)$ into a directional environment map $L(\boldsymbol\omega_i)$ and a light visibility term $f_v(\mathbf{x}, \omega_i)$,
\begin{equation}
L_i(\boldsymbol\omega_i,\mathbf{x}) = f_v(\boldsymbol\omega_i,\mathbf{x})L(\boldsymbol\omega_i)
\label{equ:light_visibility}
\end{equation}
A straightforward way to estimate the NReF is by simply inserting equ.~\ref{equ:rendering} into equ.~\ref{equ:nerf_render} and minimizing the render loss with image observations.
However, simultaneously estimating all components of NReF from scratch is extremely hard and unstable due to its ill-posedness nature, even under known illumination conditions~\cite{nerv2021}. Fortunately, the NeRF has already decomposed geometry information to some extent and we can extract an initial surface $\mathbf{S}$ from it. Given this, the rendering loss $\mathcal{R}(\mathbf{r})$ can be then greatly reduced to first query the surface point $\mathbf{x}_s$ and then evaluate equ.~\ref{equ:rendering} on it:
\begin{equation}
\mathcal{R}(\mathbf{r}) = \mathcal{R}(\boldsymbol\omega_v, \mathbf{x}_s) = \lVert I(\mathbf{r}) - L_o(\boldsymbol\omega_v,\mathbf{x}_s)\rVert^2_2
\label{equ:render_loss}
\end{equation}

The numerical method for approximating the integral of equation~\ref{equ:rendering} plays a crucial role during the optimization. Previous neural reflection field estimation method~(\cite{zhang2021nerfactor}, \cite{nerv2021}) approximate the integration with a pre-defined equirectangular map of lighting directions. However, we argue that this simple strategy is far from an optimal one~(\cite{veach1995optimally}). In particular, this sampling strategy is not only biased but also gives significant noisy results with an affordable amount of samples during training. Naively increasing number of samples leads to unacceptable memory and time cost. We address this issue by following the standard importance sampling strategy~\cite{veach1995optimally} in physical-based rendering field. The importance sampling directions are calculated based on the material roughness properties.
%Actually, those are fundamental techniques in physical based rendering, and we found it crucial to reduce both forward estimation variance and backward gradient variance. 

\subsection{Extracting geometry priors from NeRF}
\label{sec:method:surface_extraction}
\begin{figure}[!ht]
    \centering
    \includegraphics[width=0.9\linewidth]{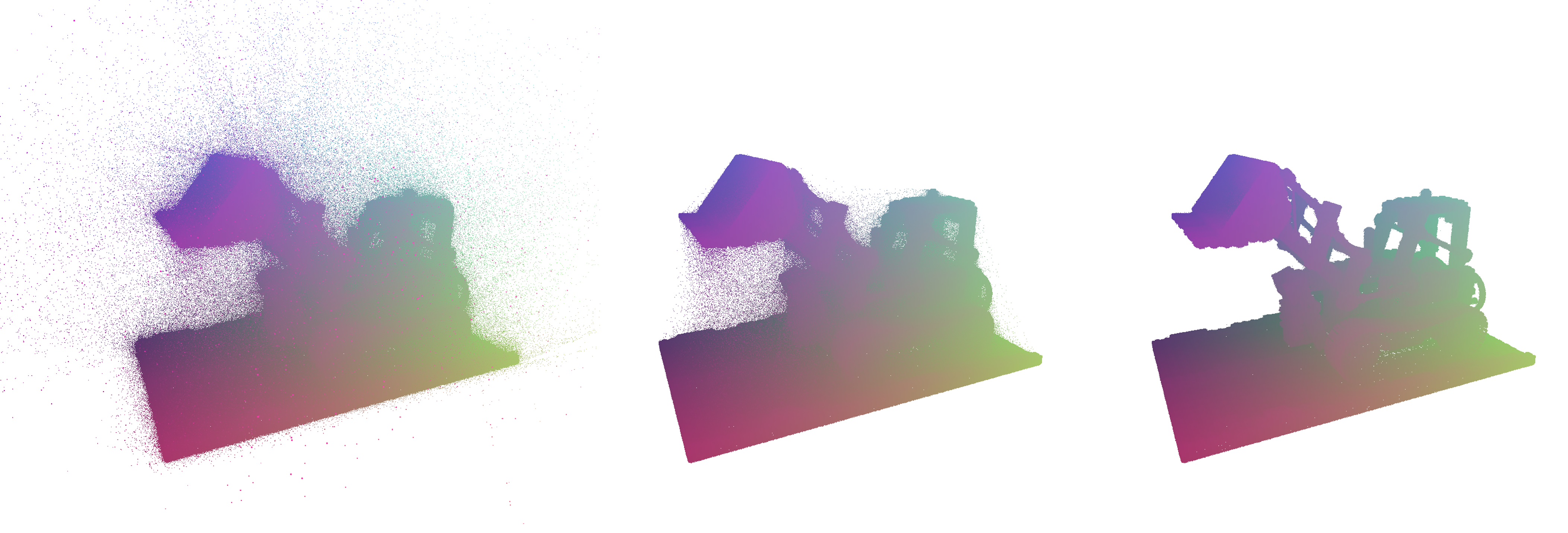}
    \caption{The improvement of surface extraction. Left: surface extraction result with equation~\ref{equ:surface_wrong} produces many erratic scattered points.
    Middle: A slightly improved version of  equation~\ref{equ:surface_wrong} by normalizing weight still produces scattered points.
    Right: Our surface extraction method with equation~\ref{equ:raymarching} removes almost all outlier points.} 
%    From left to right, surface extracted using Equ.~\ref{equ:surface_wrong}, an improved version by normalizing the weight and ours method(Equ.~\ref{equ:raymarching}.}
    \label{fig:shape_improvement}
\end{figure}
\noindent{{\textbf{Surface points and normals}}~
The original NeRF suggests extracting surface points along a single ray $\mathbf{r}(t)$ with its expected termination:
\begin{equation}
\mathbf{x}_s=\int_{0}^\infty \frac{\partial T(\mathbf{r}(t))}{\partial t}\mathbf{r}(t)dt
\label{equ:surface_wrong}
\end{equation}
The surface normal at $\mathbf{x}_s$ can be computed as the negative normalized gradient of NeRF's density output $\sigma(t)$ w.r.t point positions via auto-differentiation~\cite{nerv2021}\cite{zhang2021nerfactor}.
In practice, however, we observe that the surface and normal derived from equ.~\ref{equ:surface_wrong} is usually noisy and erratic, as shown in fig~\ref{fig:shape_improvement} and fig~\ref{fig:normal_improvement}.
The reason is that the density field from NeRF actually tends to `fake' glossy surfaces by creating two or more small layers, thus naively blend them along ray directions will create `fake' floating points. The detailed analysis of failure cases of equ.~\ref{equ:surface_wrong} are given in the supplementary material.
To alleviate this, we employ an empirical but effective strategy by simply finding the point $\mathbf{x}_s$ on the ray that satisfies
\begin{equation}
T(\mathbf{r}(s))=\frac{T(\mathbf{r}(t_n))+T(\mathbf{r}(t_f))}{2}
\label{equ:raymarching}
\end{equation}
where $t_n$ and $t_f$ are the ray-tracing bound. Unlike equ~\ref{equ:surface_wrong} which spreads floating points along the ray, extracting points with equ~\ref{equ:raymarching} will force the points distributed on one of the surface layers.
As $T(\cdot)$ is a monotonic function by definition, there is always a unique solution for equ.~\ref{equ:raymarching} and we found it works well in practice, as shown in fig~\ref{fig:shape_improvement}.
Given the surface point $\mathbf{x}_s$, we extract its normal directions by averaging the density gradient around a small region centered at  $\mathbf{x}_s$ weighted by its density value to further reduce the normal noise.\\

\noindent{{\textbf{Light visibility}}~
The light visibility at surface point $\mathbf{x}_s$ can be calculated as integrating the $T(\cdot)$ defined in equ.\ref{equ:transmit} on the normal-directed hemisphere.
Since we have built an octree to store the density field on the surface points, we again utilize it to compute the integration dynamically with importance sampling strategy during the optimization.
%To compute the integral, we found that a static sample strategy used on the lighting directions in previous work~\cite{zhang2021nerfactor} are far from an optimal sampling strategy, giving significant noisy results with the same affordable amount of samples during training.
%As the optimal importance sampling strategy is dynamic w.r.t. surface normals and material roughness properties, we construct an octree~\cite{yu2021plenoctrees} on top of the extracted surface points to store the NeRF density field.
Note that the octree not only supports light visibility query on the fly, but also enables efficient depth estimation during the rendering. \\

\noindent{{\textbf{NeRF commitment}}~~During the decomposing process, the surface normal and light visibility of NReF will be refined by the render loss. Yet, the predicted normal and light visibility should not derive too much to that of NeRF. Thus, we add a NeRF commitment loss to constrain the optimized normal and visibility close to NeRF on the extracted surface points,
\begin{equation}
    \label{equ:commitment_loss}
    \mathcal{C}_{f_\theta}(\mathbf{x_s}) = \lVert f_\theta(\mathbf{x}_s) - F(\mathbf{x}_s) \rVert^2_2
\end{equation}
where $f_\theta(\cdot)$ and $F(\cdot)$ denote corresponding components (normal or visibility) of NReF and NeRF respectively, $\theta$ indicates the dependence of network parameters of NReF. 

%The behind reason is that the original shape inducted from NeRF often creates small, double-layered translucent region with vacuumed density in-between, which violates the assumption of opaque surfaces.
%\input{figures/fig_method_visualize_T}
%This phenomenon itself, and how it affects the normal induction, is illustrated in figure~\ref{fig:method_visualize_T}. An theoretical opaque surface should have a step-wise transmittance function along ray direction (red curve) where the step-change point corresponds to surface point.
%In practice, transmittance function trained by NeRF usually have with two "small step" close to each other, corresponds to a double-layered translucent surface. Applying equation~\ref{equ:surface_wrong} in such case will hit a wrong approximation of surface point with small, unstable gradient and thus leading to noisy, erratic normal.

\subsection{Enforcing Smooth and Parsimony}
\label{sec:method:priors}
For estimating reflectance under unknown illumination, other priors are necessary. We employ two well-known prior knowledge from the intrinsic decomposition literature~\cite{BarronTPAMI2015}: the predicted normal, visibility and BRDF should be locally smooth, and the albedo color should be globally sparse.
These priors can be unified into one single formulation as applying filter operations on the surface points (the derivation and normalizing term have been dropped for simplification):
\begin{equation}
    \label{equ:prior_filter}
     f_\mathcal{P}(\mathbf{x}_i) = \sum_{j \neq i}k(\mathbf{v}_i,\mathbf{v}_j)f_\theta(\mathbf{x}_j)
\end{equation}
where $k(\mathbf{v}_i,\mathbf{v}_j)$ denotes the weight from $\mathbf{x}_j$ to $\mathbf{x}_i$, based on the similarity of a high-dimensional vector $\mathbf{v}$ defined on the surface. The form of $\mathbf{v}$ is related to different prior types. Specifically, for local smoothness term, $\mathbf{v}$ is defined as the bilateral kernel weight:
\begin{align}
    \mathbf{v}_i &= (\mathbf{x}_i, f_\theta(\mathbf{x}_i))\\
    k(\mathbf{v}_i,\mathbf{v}_j) &= \exp(-|\mathbf{v}_i - \mathbf{v}_j)|^2)\\
    &= \exp(-|\mathbf{x}_i - \mathbf{x}_j|^2) \exp(-|f_\theta(\mathbf{x}_i)- f_\theta(\mathbf{x}_j)|^2)
\end{align}
For global sparsity (also known as parsimony in~\cite{BarronTPAMI2015}), $\mathbf{v}$ contains only the albedo color, i.e., $\mathbf{v}_i=f_\theta(\mathbf{x}_i)$.
During training, priors are enforced by first applying filter operations with equ.~\ref{equ:prior_filter}, then minimizing the difference between the NReF $f_\theta(\cdot)$ and its filtered version $f_\mathcal{P}(\cdot)$:
\begin{align}
    \label{equ:prior_loss}
    \mathcal{P}_{f}(\mathbf{x}_i) &= \lVert f_\mathcal{P}(\mathbf{x}_i) - f_\theta(\mathbf{x}_i)\rVert\\
    &= \lVert \sum_j k(\mathbf{v}_i,\mathbf{v}_j)(f_\theta(\mathbf{x}_j) - f_\theta(\mathbf{x}_i))\rVert\\
    \label{equ:prior_loss_approximate}
    &\approx \sum_j k(\mathbf{v}_i,\mathbf{v}_j)\lVert f_\theta(\mathbf{x}_j)-  f_\theta(\mathbf{x}_i)\rVert
\end{align}
\noindent{\textbf{Computing priors with Gaussian KD-Tree}}~
In practice, we calculate the prior terms using the Gaussian KD-Tree~\cite{adams2009gaussian}.
A high-dimensional KD-Tree is constructed on point set of vector $\mathbf{v}$. 
%We will down-sample the surface points when necessary for efficiency.
%Notice that downsampling is actually encouraging smoothness.
To calculate the prior loss, instead of directly apply filtering and storing the filtered values $f_\mathcal{P}$, we approximate it stochastically by using an importance sampling of $\mathbf{x}_j$ proportionally to $k(\mathbf{v}_i,\mathbf{v}_j)$ in each mini-batch (see equ.~\ref{equ:prior_loss_approximate}), and asynchronously updating the KD-tree to reflect changes of $\mathbf{v}$ every epoch.
%Notice that the KD-Tree is dynamic and need to be updated to reflect changes of $\mathbf{v}$. In our implementation, we re-build the KD-Tree every epoch.
For parsimony term, we further reduce the computational cost by employ a K-means operation on the albedo colors and randomly keep several points in each cluster as the candidate $\mathbf{x}_j$.
%For parsimony term, naively using the Gaussian KD-Tree may resulting in too many neighbors $\mathcal{x}_j$, instead, we employ a K-means operation on the albedo colors. For each cluster we randomly keep several points as the candidate $\mathbf{x}_j$.
%for each $\mathbf{x}_i$, we need to additionally sample other points $\mathbf{x}_j$ on surface. 
%For each $\mathbf{x}_i$, we can draw $\mathbf{x}_j$ proportionally to $k(\mathbf{x}_i,\mathbf{x}_j)$ by using a importance nearest neighbor sampling. Notice that if we set the spatial derivation to infinity, this function will achieve global sparsity. However, we found that this will lead to too many sampling points. For smoothness,the Gaussian KD-Tree is very efficent because the spatial distance will be the dominant part, while for sparsity, we use a modifed version by applying K-means on the albedo and randomly keep only several of samples. This way, each sample will have a limited sparsity neighborhood, 

%, and accumulate the rendering loss using Monte Carlo estimation as 
%\begin{equation}    f_a + \frac{1}{N_s} \sum ...\end{equation}

%\input{figures/fig_method_visibility}
\subsection{Multi-stage Optimization}
Per equation~\ref{equ:render_loss},~\ref{equ:commitment_loss} and~\ref{equ:prior_loss}, the final loss used to decompose NReF from NeRF is:
\begin{equation}
    \label{equ:final_loss}
    \mathcal{L} = \lambda_r\mathcal{R} + \lambda_c\mathcal{C} + \lambda_p\mathcal{P} 
\end{equation}
To avoid optimizing a too complicated target space, we split the NReF decomposition into multiple sub-stages as follows (more details in the supp. material):\\

\label{sec:method:nref_decomposition}
\noindent{\textbf{Pre-training NeRF}}~
We use the same network structure and follows the training strategy from~\cite{mildenhall2020nerf} for pre-training NeRF. We train NeRF for 2000 epochs with randomly sampling 1024 camera rays for each image per mini-batch.\\

% \paragraph{\textbf{Extracting geometry priors}} Given a pre-trained NeRF, we firstly extract the surface points and normals using Equ.~\ref{equ:raymarching}.
% The extracted point cloud contains $500K$ points in average. We then build a KD-Tree and an octree on them. Each point will also be assigned an albedo color as well as a roughness value, and they will be initialized at the final joint optimization stage discussed later.\\

\noindent{\textbf{Training normal and visibility}}~
Given a pre-trained NeRF, we firstly extract the surface points and normal using equ.~\ref{equ:raymarching}. An octree is constructed on the surface points and the light visibility is computed on the fly during training.
We train the normal and visibility component for 100 epochs. 
During training we filter out few points that still have erratic normal directions our refined surface extraction by discarding their commitment loss and additionally add a visibility loss to predicted normal w.r.t. its view direction. \\
%For the visibility part, we sample 64 diffuse uniformly and 128 specular using random roughness for each iteration.\\

\noindent{\textbf{Joint optimization}}~
After the normal and visibility training, we have had a quite reasonable geometry initialization for NReF.
To prevent the uninitialized material properties and env. lighting ruin out the geometry at the beginning, we apply a warm-up stage during which we use large geometry commitment weight $\lambda_c$ and small smoothness and parsimony weight $\lambda_p$.
We warm-up the training for 100 epochs and then joint train all components with equ.~\ref{equ:final_loss} for another 200 epochs with all terms properly applied.\\
%During training, we follow a classical importance sampling technique described in~\cite{mitchell2007gpu} for evaluating rendering equation (Equ~\ref{equ:rendering}), with 64 rays uniformly sampled for the diffuse part and 128 rays sampled for the specular part.\\
%We do not add additional priors on roughness/BRDF component as opposed to~\cite{zhang2021nerfactor}. We argue that their requirement of using a pre-trained BRDF model as an additional prior is due to its improper sampling strategy which cannot handle glossy surfaces. 

\noindent{\textbf{Implementation details}}~
%The optimization process is implemented with PyTorch~\cite{paszke2019pytorch}.
We model the BRDF $f_r(\cdot)$ as a lambertian diffuse part with albedo color $f_a(\mathbf{x})\in R^3$ plus a specular part with GGX Microfacet model~\cite{walter2007microfacet} that is controlled with roughness parameter $f_s(\mathbf{x})\in[0,1]$.
The albedo $f_a(\mathbf{x})$, roughness $f_s(\mathbf{x})$, normal $f_n(\mathbf{x})$, and visibility $f_v(\mathbf{x}, \omega_i)$ are all parameterized with MLP network of 4 layers. The environmental map $L(\boldsymbol\omega_i)$ is represented with a cube map texture. We build a mipmap over the cubemap and sample the corresponding mipmap level according to the PDF of sampled light directions~\cite{mitchell2007gpu} using soft rasterization~\cite{Laine2020diffrast}. 
For computing equ.~\ref{equ:rendering}, we sample 128 rays w.r.t the estimated roughness and 64 rays with uniform sampling.
We set $\lambda_r=1.0$; the weight $\lambda_c$ is set to 0.5 during the warm-up training, and reduced to 0.1 in the final joint training stage. The weight $\lambda_p$ for albedo/roughness/shape smoothness is set to 0.5/0.01/0.1, and 0.1/0.005 for albedo/roughness parsimony.
\\

\noindent{\textbf{Computational cost}}~
The whole training can be conducted on a single NVidia Tesla V100 GPU. The total training time for $512\times 512$ resolution with $100$ views is approximately 15 hours, with 14 hours for NeRF pre-training, 15 minutes for training normal and visibility, and another 30 minutes for joint optimization.
For inference, rendering one image of $512\times512$ takes around 8 seconds with a typical importance sampling setup of 64 diffuse samples and 128 specular samples.

\section{Experiments}
\label{sec:expr}
To validate our proposed method, we first perform ablation studies on our revised geometry extraction method and different prior terms enabled by octree and Gaussian KD-tree (sec.~\ref{sec:expr:ablation}).
We also perform comparisons against related methods and show our advantage (sec.~\ref{sec:expr:comparsions}).
Finally, we show more results on real data and demonstrate manipulation applications enabled by NReF (sec.~\ref{sec:expr:real_capture}).\\

\noindent{\textbf{Datasets}}~~The ablation studies and comparison with~\cite{zhang2021nerfactor} and \cite{nerv2021} is evaluated on the synthetic Blender scenes released by Mildenhall et al. (\cite{mildenhall2020nerf}). In~\cite{zhang2021nerfactor} the author re-render the synthetic Blender scenes with their own illumination conditions. We compare our results using their rendering setup for a fair comparison. The comparison with~\cite{Xia:2016:RSS} is evaluated on their real captured dataset. Results on other real data is generated from mobile phone captured data released by~\cite{mildenhall2020nerf}.
% \paragraph{Metrics}
% For quantitative evaluations, we report Peak Signal-to-Noise Ratio (PSNR), Structural Similarity Index Measure (SSIM)~\cite{wang2004image} for albedo and rendered images and average angle difference (in degrees) for normals. Quantitative comparison of rouhness is not directly applicable as different BRDF model is employed for different methods and we refer to novel light re-render as an indirect measurement of roughness estimation.
\subsection{Ablation Studies}
\label{sec:expr:ablation}
In this section we ablation each component in our optimization pipeline that contributes to the final high-quality reflectance field results.\\

\noindent{\textbf{Geometric-aware smoothness term}}~~
A significant advantage of our method is the geometric-aware smoothness enabled by bilateral normal filtering with Gaussian KD-tree.
We validate the gain from this advantage in fig.~\ref{fig:smoothness_improvement} by removing the smoothness term, or replacing it with a naive smoothness term in euclidean space (i.e., similar to~\cite{zhang2021nerfactor}). Without geometric-aware smoothness, the geometry features tend to be either too noisy (w/o smoothness at all) or smoothed out together with noise.\\
\begin{figure}[!ht]
    \centering
    \includegraphics[width=0.8\linewidth]{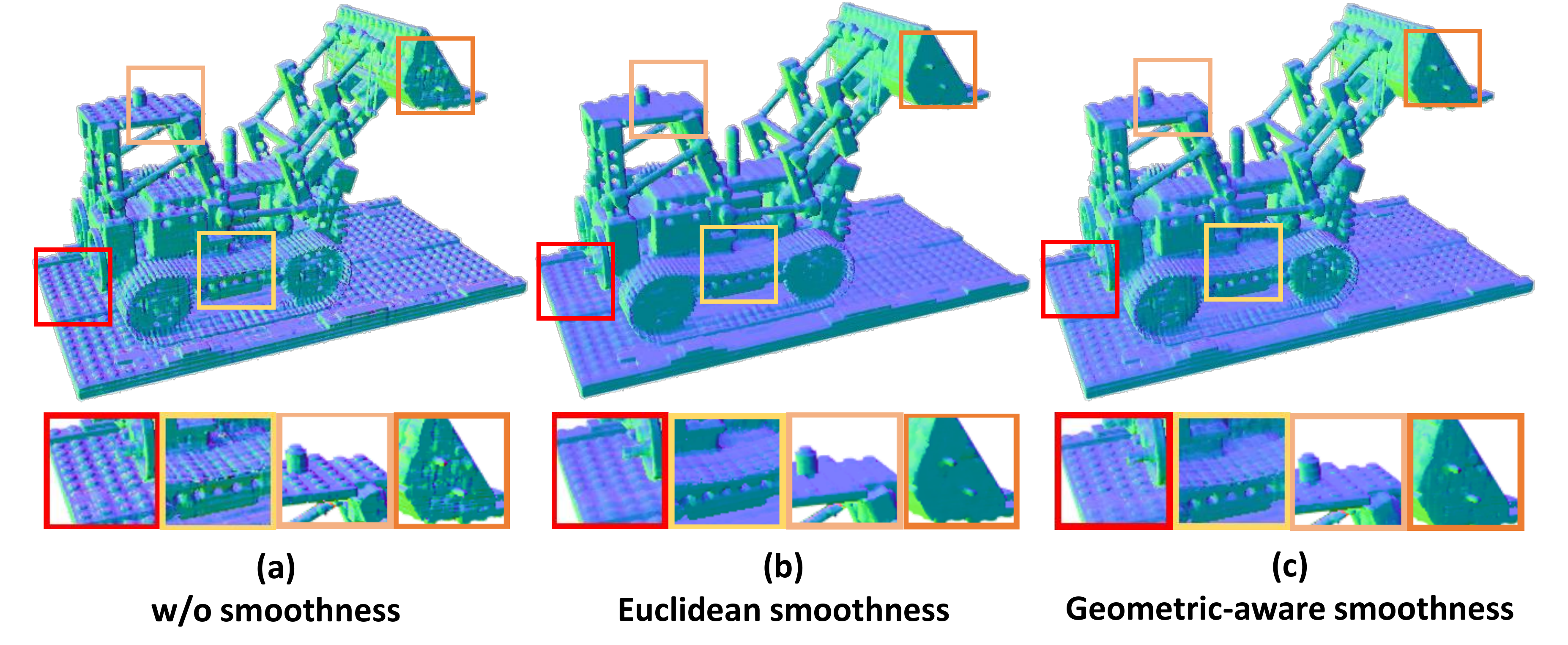}
    \caption{The effect of smoothness term. (a) without any smoothness constraint, the trained normal is sensitive to noisy points (e.g., the tiny lego bumps, highlighted in red) (b) directly adding smoothness in euclidean space remove most noise; however, it also tends to smooth out geometry details. (c) adding geometric-aware smoothness removes noisy normal while preserves shape details.}
    \label{fig:smoothness_improvement}
\end{figure}

\noindent{\textbf{Global parsimony term}}~
\begin{figure}[!ht]
    \centering
    \includegraphics[width=0.9\linewidth]{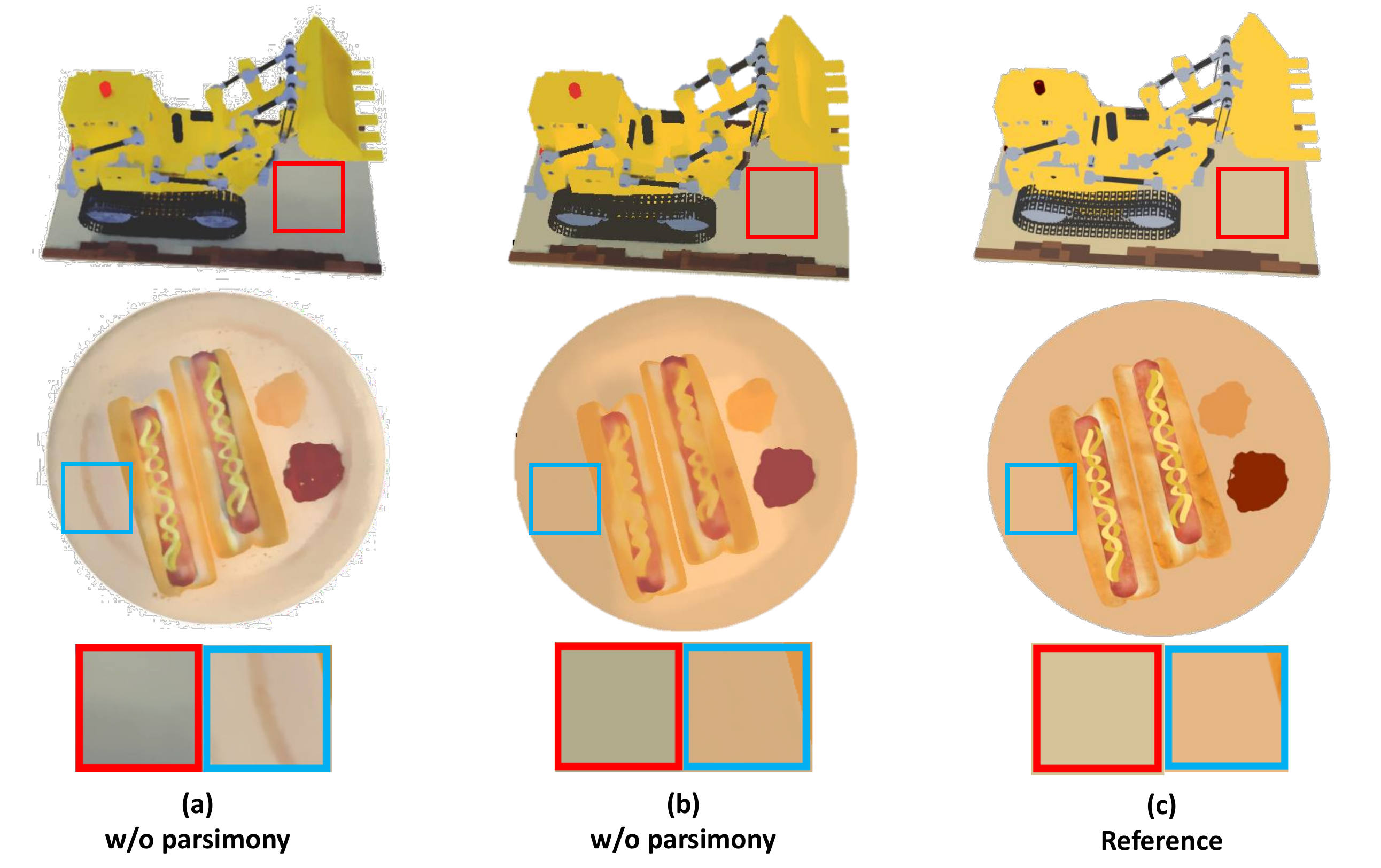}
    \caption{The effect of parsimony term. (a) the shading effect is hard to disentangled from albedo without a parsimony prior.
    (b) the parsimony term enforces a piecewise-constant albedo and remove most shading. Note for the bottom case our method (b) fails to completely produce a single color for the plate due to strong inter-reflection that cannot be handled in our formulation. Yet, we still remove most shadow caused by shading compared to (a).}
    \label{fig:parsimony}
\end{figure}
Our method also enables global prior terms that cannot be applied in previous works~\cite{nerv2021},\cite{zhang2021nerfactor}.
We show the benefit of our global parsimony term in fig.~\ref{fig:parsimony}. The global parsimony provides a strong color-sparsity constraint on the albedo field and prevents incorrect shading effects baking in.\\

\noindent{\textbf{Improvements over NeRF normals}}~
\begin{figure}[!ht]
    \centering
    \includegraphics[width=0.9\linewidth]{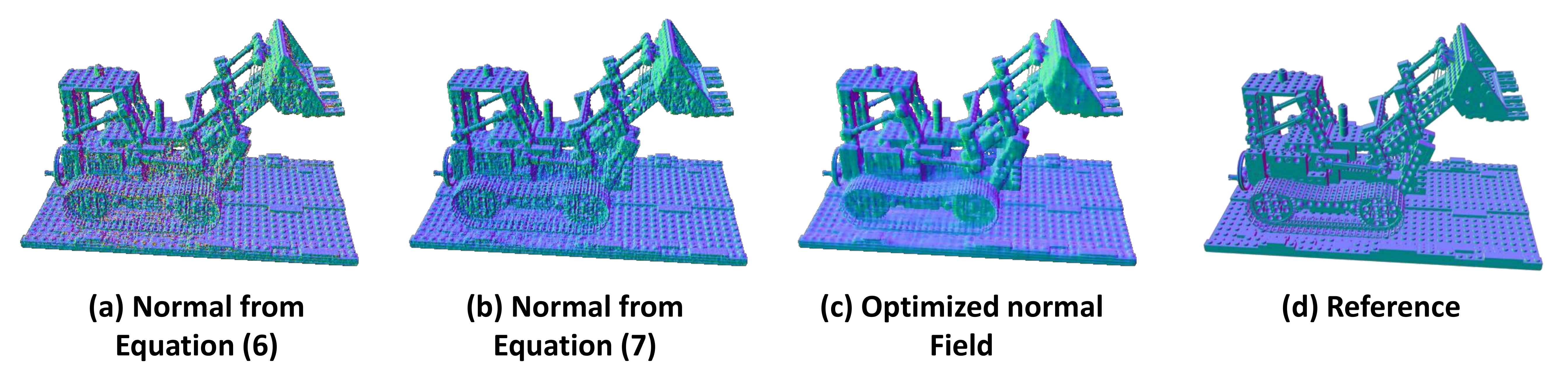}
    \caption{The improvement over NeRF normal. (a) Initial normal extracted with Equation~\ref{equ:surface_wrong} is extremely noisy and erratic such as the track region. (b) Initial normal extracted with Equation~\ref{equ:raymarching} improves the normal and provides a good initialization point. (c) Our optimization further improve the normal with significant noise reduction while preserves geometry features. The reference normal is shown in (d).}
    \label{fig:normal_improvement}
\end{figure}
We compare and validate our effects of better normal initialization as well as normal optimization results in
fig.~\ref{fig:normal_improvement}. The initial normal extracted with equ.~\ref{equ:surface_wrong} is extremely noisy (e.g., the track region). Our method extracts a cleaner initialization with equ.~\ref{equ:raymarching} and the results are further improved after the optimization.\\

%\noindent{\textbf{Effect of light sampling}}~~One evidence is that the light we have estimated is much close to the groundtruth than the base line method.(including NeRFactor).
%\input{figures/fig_application}
\subsection{Comparisons}
\label{sec:expr:comparsions}
\begin{figure}[!ht]
    \centering
    \includegraphics[width=0.9\linewidth]{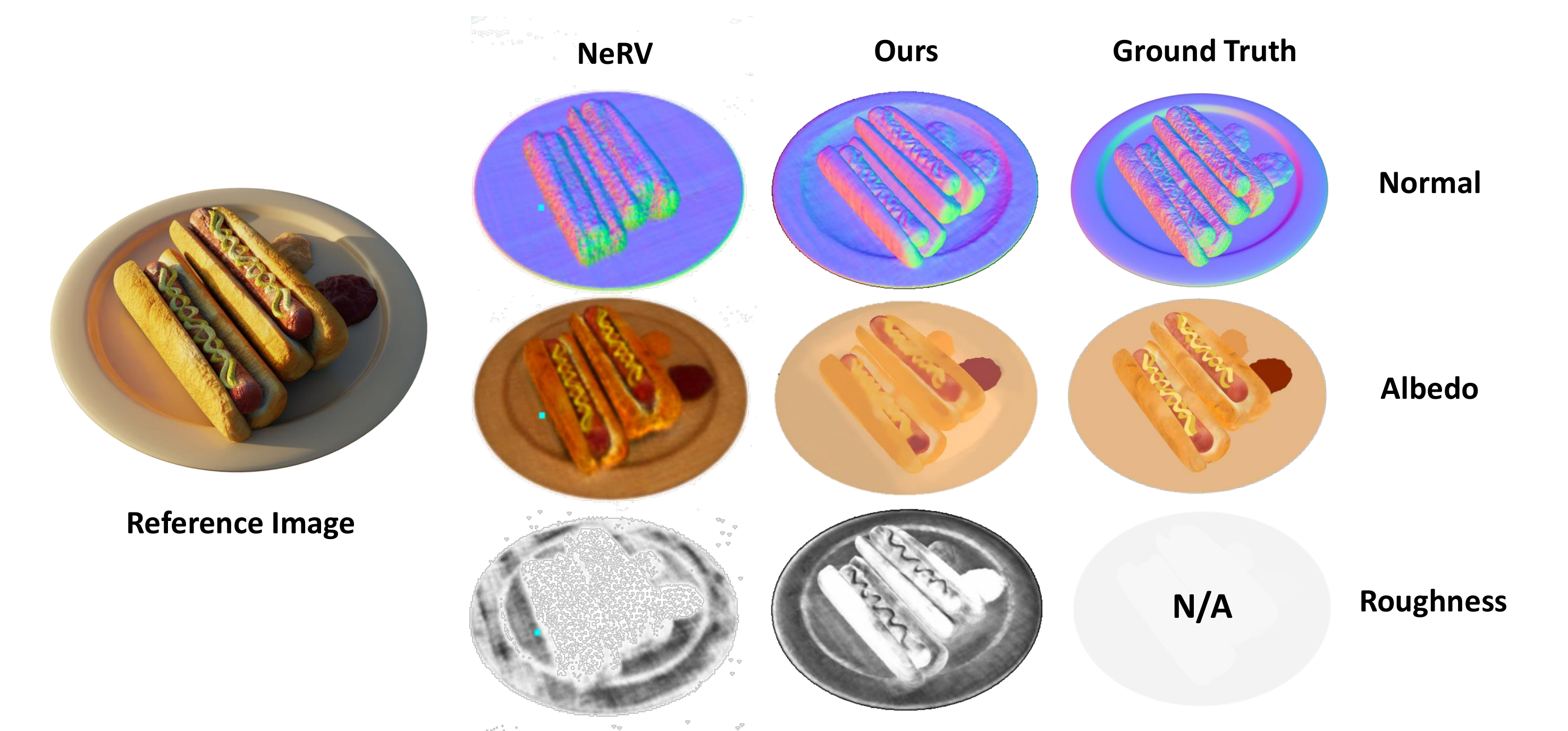}
    \caption{Comparison with NeRV~\cite{nerv2021}. NeRV (left) produce incorrect normals and noisy roughness on the plate part, with a large ratio of shading baked into albedo. Our result (middle) produces correct normal for the plate, remove most shading from albedo and produces a reasonable roughness with glossy material for the plate and rough material for the hotdog. One example view of the input image is shown on the left for reference to illustrate the strong inter-reflection in this data.}
    \label{fig:comparsion_nerv}
\end{figure}
\begin{figure*}[!ht]
    \centering
    \includegraphics[width=0.65\linewidth]{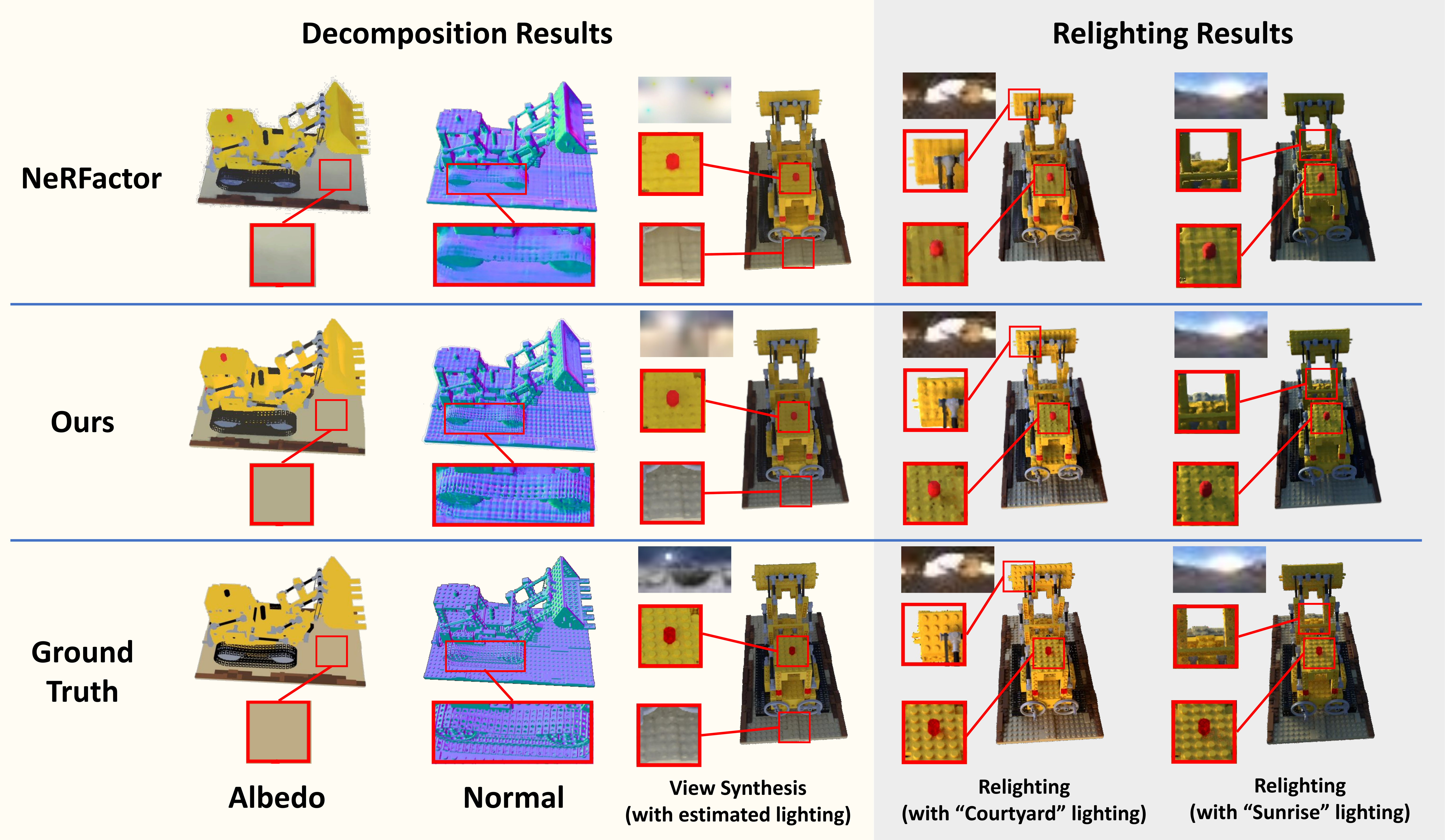}
    \caption{Compared with NeRFactor~\cite{zhang2021nerfactor}, our method estimates a cleaner albedo with less shading bake in, a more detailed normal without introducing noise, and more plausible lighting conditions. Our method also renders novel images that better reproduces geometry details and lighting effects. The lighting estimated from input and used for re-rendering are shown on the left-top of corresponding images.}
    \label{fig:comparsion}
\end{figure*}
%\begin{table}[ht!]
%    \centering
%    \resizebox{0.45\textwidth}{!}{
%    \begin{tabular}{c|c|c|c|c}
%         &  Albedo & Normal & Novel View & Relighting \\
%    \hline
%    NeRFactor &  28.71   &   22.13  & 32.54 & 26.63 \\
%    Ours & 28.71 & 22.13 & 32.54 & 26.63\\
%    \hline
%    \end{tabular}
%    }
%    \caption{Quantitative comparison with NeRFactor~\cite{zhang2021nerfactor}. For normal we compare the average angle error in degrees; for albedo and synthesised image we compare PSNR. The roughness are not applicable for quantitative comparisons because NeRFactor employs a different BRDF model.}
%    \label{tab:comparsion}
%\end{table}
\begin{table*}[ht!]
    \centering
    %\resizebox{\linewidth}{!}
    \begin{tabular}{c|c|c|c|c}
    \hline
    Method & Normal (Degree)$\downarrow$ & Albedo (PSNR)$\uparrow$ & View Synthesis (PSNR)$\uparrow$ & Relighting (PSNR)$\uparrow$ \\
    \hline
    NeRFactor\cite{zhang2021nerfactor} & 22.1327 & 28.7099 & 32.5362 & 23.6206\\
    NeRFactor* & \textbf{29.0603} & 22.1496 & 24.8610 & 19.0691 \\
    Ours & 30.1381 & \textbf{24.0959} & \textbf{27.2362} & \textbf{20.5153} \\
    \hline
     \end{tabular}
    \caption{Quantitative comparison with NeRFactor. We found that the original implementation of NeRFactor computes the PSNR in grayscale color space and did not remove the black backgrounds, leading to lower errors on all scenes. For fairness, we list both their reported values (first row) and re-computed values in RGB with only object foreground (second row). 
    Our method (bottom row) exhibits a significant PSNR improvement (1.95db/2.38db/1.45db) for albedo and novel synthesized image with refined metric. NeRFactor have a slightly lower normal error (1.07$^{\circ}$) than us, but visually we found our method produce more detailed results on normal map~(fig~\ref{fig:comparsion}).}
    \label{tab:compare_nerfactor}
\end{table*}
\noindent{\textbf{Comparisons with other neural reflectance field}}~~We compare our methods with two methods that share a similar setup: NeRFactor~\cite{zhang2021nerfactor} and NeRV~\cite{nerv2021}.
%The qualitative results are shown in fig.~\ref{fig:comparsion_nerv} and fig.~\ref{fig:comparsion}.
NeRV~\cite{nerv2021} directly train everything from scratch without employing a NeRF pre-training stage, leading to a too complicated optimization problem that often fall into local minima with less plausible visual quality, as shown in fig.~\ref{fig:comparsion_nerv}.
NeRFactor~\cite{zhang2021nerfactor} formulates the problem similar to us by first pre-training a NeRF and then conducting decomposition. However, without tree-based structures as support, their method is optimized without a dedicated consideration on surface point extraction, suitable prior terms, and sampling strategies.
Thus, it suffers both smoothness and parsimony issues as discussed in sec.~\ref{sec:expr:ablation}, producing over-smoothed normal field as well as less cleaned albedo field, and often degenerates to a near-diffuse decomposition result. 
Our method address all the above issues and outperforms NeRFactor, both quantitatively (tab.~\ref{tab:compare_nerfactor}) and qualitatively (fig.~\ref{fig:comparsion}).\\

\noindent{\textbf{Comparison with non-neural method}}~~We compare our method on real captured data with a high-quality non-deep learning method~\cite{Xia:2016:RSS} that reconstruct shape, appearance and lighting from image sequences. Fig~\ref{fig:comparsion_xfm} shows a qualitative comparison result.
As reported in~\cite{Xia:2016:RSS}, their results were generated with a 20 node PC cluster using a total of 1243 captured images. Our method reduced the number of images needed (500 images for this experiment) and can be conducted on a single PC with one GPU card. Overall, the quality of our results are similar to~\cite{Xia:2016:RSS} with significantly reduced number of inputs and computational cost.
\begin{figure*}[!ht]
    \centering
    \includegraphics[width=0.65\linewidth]{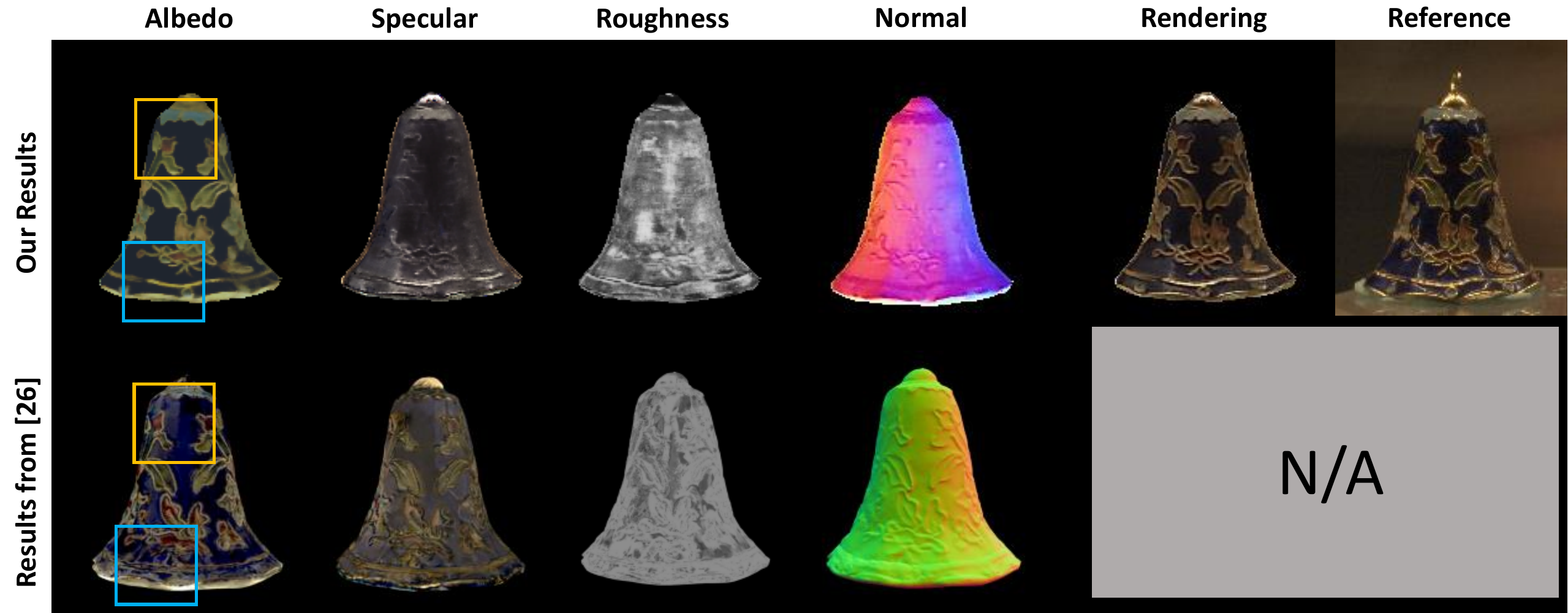}
    \caption{Qualitative comparison with~\cite{Xia:2016:RSS} on the real captured "Cloisonne Bell" data from~\cite{Xia:2016:RSS}. Note that~\cite{Xia:2016:RSS} used a different BRDF model thus the absolute value of roughness and specular are not comparable, and~\cite{Xia:2016:RSS} use a different color maps for normal visualization.
    Overall, our method (top row) produces a cleaner albedo (highlighted with yellow rect.) with less specular baked-in (highlighted with blue rect.) than~\cite{Xia:2016:RSS} (bottom row).
    \cite{Xia:2016:RSS} produces visually cleaner roughness than ours. Yet, both method produce a visually plausible specular components and our rendering results are plausible comparing with the reference image.}
    \label{fig:comparsion_xfm}
\end{figure*}
\begin{figure*}[!ht]
    \centering
    \includegraphics[width=0.65\linewidth]{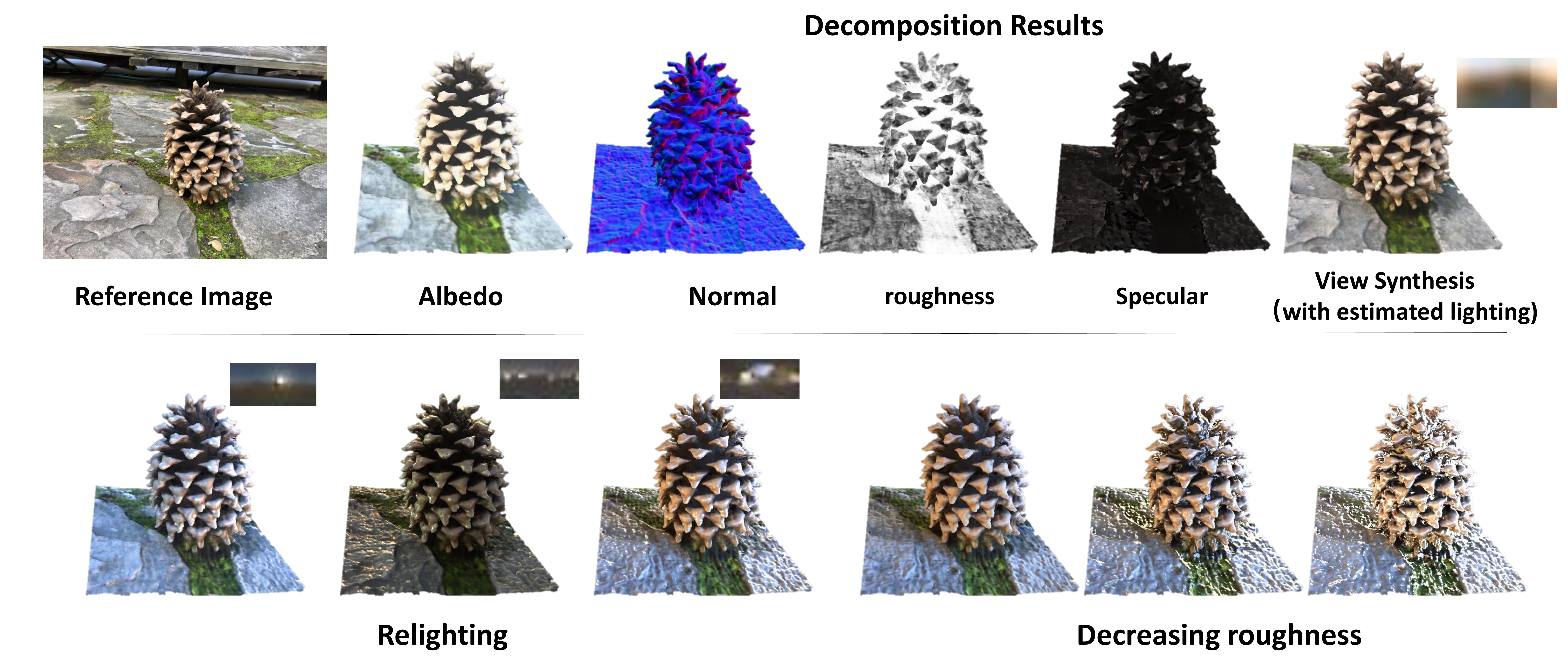}
    \caption{NReF estimation on real captured data. Left: One of the input views for reference. Upper row: decomposed albedo, normal and novel view synthesis with estimated lighting on the right. Lower row: relighting and changing roughness.}
    \label{fig:real}
\end{figure*}
\subsection{Results and Applications}
\label{sec:expr:real_capture}
We further demonstrate our method’s ability to estimate NReF as well as its applications for manipulation tasks on real-world captured multi-view images. Fig~\ref{fig:real} shows our decomposed NReF components, view synthesis, relighting, and material editing results. Our method estimates plausible reflectance field decomposition and enables photo-realistic editing results. Additional results, including videos of relighting and view synthesizing, are given in the supp. material.

%\input{figures/fig_failure}

%\subsection{Applications}
%\label{sec:expr:applications}
%Finally, we demonstrate the editing applications enabled by our NReF estimation results. Fig.~\ref{fig:application} shows our results on relighting and material editing. Thanks to our high-quality decomposition of shape, appearance and lighting, we are able to get plausible, photo-realistic results.

%\textbf{Ablation}:
%\begin{itemize}
%    \item No smoothness, Naive Euclidean smoothness, Geometric-Aware Smoothness
%    \item no parsimony
%    \item NeRF shape (old) / NeRF shape (new) / Our optimized shape
%\end{itemize}

%\textbf{Comparisons}:
%\begin{itemize}
%    \item NeuralFactor (quality, robustness)
%    \item NeRV (quality, The importance of multi-stage)
%\end{itemize}

%\textbf{Results}:
%\begin{itemize}
%    \item quantitative results
%    \item 1-2 Captured real data
%\end{itemize}

%\textbf{Applications}:
%\begin{itemize}
%    \item Relighting
%    \item Material Editing
%\end{itemize}

\section{Conclusion}
\label{sec:conclusion}
% Conclusion.
We presented a new method for estimating the Neural Reflectance Field (NReF) of objects that only requires a set of multi-view images under unknown lighting.
Our method is built upon a multi-stage training pipeline that decomposes a pre-trained NeRF into NReF.
The key to enable our high quality decomposition is a new method of surface point extraction from NeRF with a dynamic importance sampling strategy supported by octree, and a Gaussian KD-tree based method to apply suitable prior terms.
We demonstrated the robustness and effectiveness of our method on both synthetic and real data. Our estimated NReF can be used for manipulation tasks such as relighting and material editing.\\

% Limitations
\noindent{\textbf{Limitations}}
Our method is not without limitations.
NReF only optimizes a normal field that is defined on the surface of the origin NeRF shape. Thus NReF might fails if the NeRF shape deviates too much from ground truth.
%NReF might produce artifacts for cases with extremely thin, non-closed surfaces. This kind of geometry is hard for a surface-based representation, and the normal field is often unstable with incorrect directions.
Our method currently assumes opaque surfaces of isotropic materials without inter-reflections, and might produces artifacts for inputs with violation.
Visualizations of some typical failure cases are given in the supplemental material.\\

\noindent{\textbf{Future works}}
Avenues for future works include removing camera pose requirements as input, supporting geometry refinement during training, handling inter-reflections, and extending our method to dynamic scenes.

%%%%%%%%% REFERENCES
\clearpage
{\small
\bibliographystyle{ieee_fullname}
\bibliography{egbib}
}

\clearpage
\appendix
This appendix provide more implementation details, additional evaluations and discussions.
\section{Implementation}
\subsection{Network Structure}
%\input{figures/fig_network}
%For completeness, we reiterate the network structure of our NReF and provide a visual illustration of the structure.
\begin{figure}[!ht]
    \centering
    \includegraphics[width=0.9\linewidth]{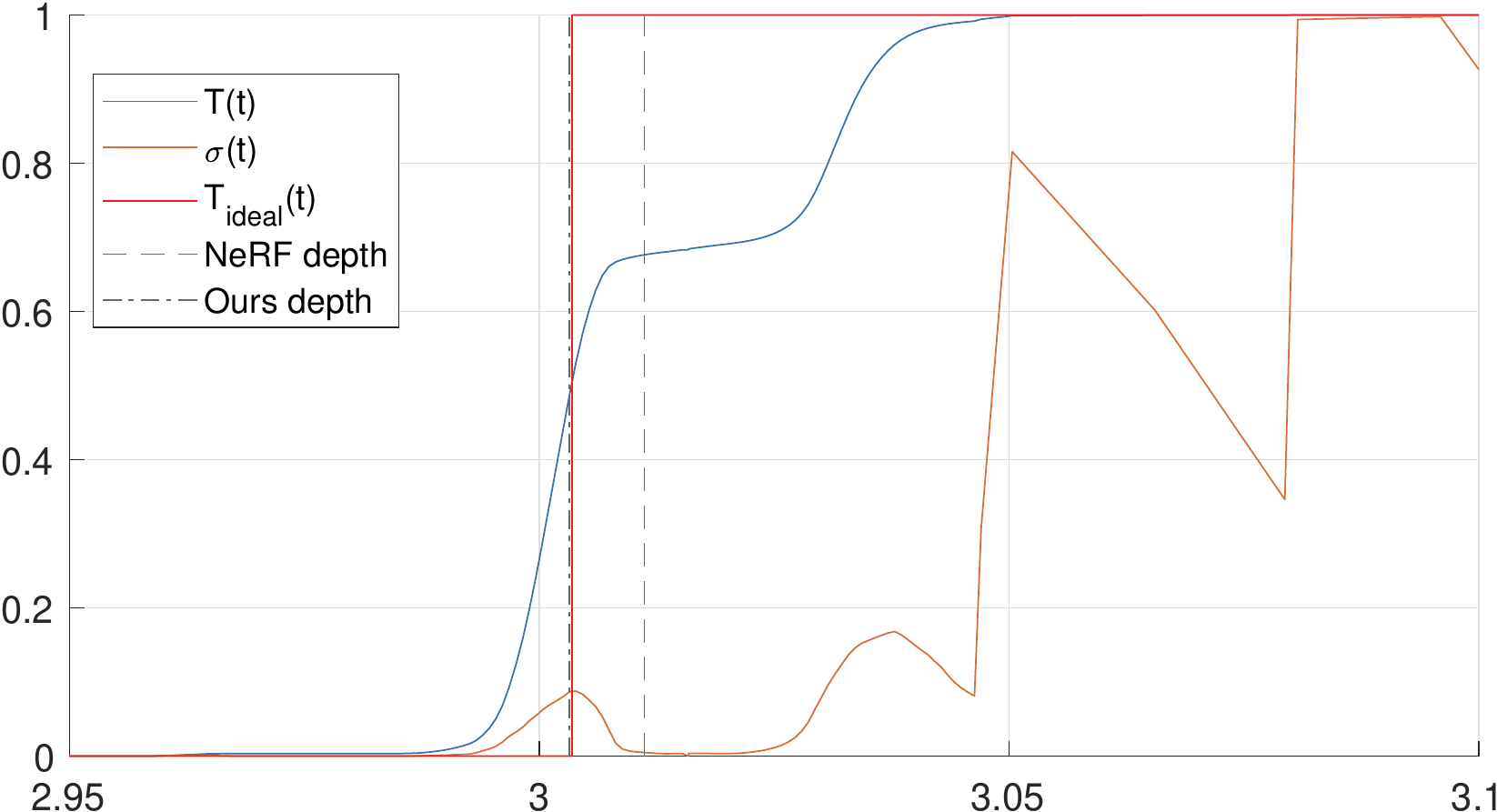}
    \caption{$T(t),\sigma(t)$ at a zooming region around the surface, along a single ray direction parameterized by $t$. The orange curve shows the density along side the ray. Notice that there is a small vacuum region between depth at 3 and 3.05. Using Equ.6 in the main paper will put the depth~(the dashed line) at this region where the gradient of the density is not well defined, i.e. nearly zero. Using our method (Equ.7) will extract a depth point with well-defined gradient~(the dot dashed line.}
    \label{fig:method_visualize_T}
\end{figure}
NReF consists of 4 sub-networks that correspond to output surface albedo color $\rho_d \in \mathbb{R}^3$, surface roughness value $\rho_r \in (0, 1)$, surface normal direction $n \in \mathbb{R}^3$, and light visibility $v \in \{0, 1\}$ respectively.
Each sub-network is modeled with an MLP network with 4 fully-connected layers with 128 hidden units and ReLU activation function.
The input of albedo, roughness, and normal is point position $\mathbf{x}\in\mathbf{S}$ on the surface. The light visibility takes both point position and incident lighting direction $\boldsymbol\omega_i$ as input by simply concatenating them. Overall, we keep our network design the same as NeRFactor~\cite{zhang2021nerfactor}, except for the BRDF component for which we use an analytical BRDF model instead of a pretrained one.
We follow the same approach of applying positional encoding to the input as the majority of NeRF-related work does.

Our pre-trained NeRF for volume density is exactly the same structure described in~\cite{mildenhall2020nerf} with 8 layers of MLP, each with a width of 256 hidden units and ReLU activation. A skip connection at the 4th layer is also included.
Figure~\ref{fig:network} details the network structures of NeRF and NReF used in this paper.\\
\begin{figure*}[!ht]
    \centering
    \includegraphics[width=\linewidth]{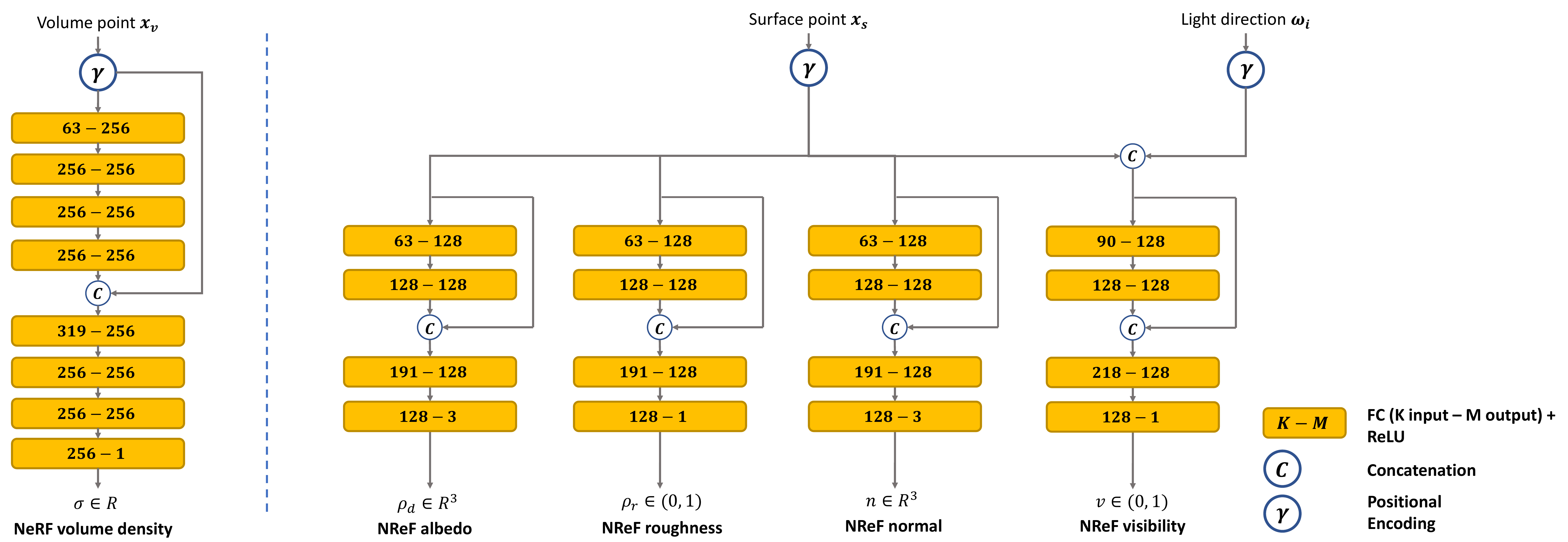}
    \caption{Our network structure for NeRF and NReF.}
    \label{fig:network}
\end{figure*}
% We use the ReLU activation function instead of Softplus function advocated by mip-NeRF~\cite{barron2021mip}, and we do not apply the mip-map position encoding strategy. As shown in figure~\ref{XX}, mip-NeRF often over-smoothes the surface normal.
\subsection{Environment lighting.}
We model the far-field lighting condition as a cube map during training.
To improve sampling efficacy and rendering quality, we construct a mip-map~\cite{mitchell2007gpu} for the cube map. The mipmap is sampled using using NVDiffrast~\cite{Laine2020diffrast}. The mipmap level for a given light ray is calculated as 
\begin{equation}
    l = \max\left(\frac{1}{2}\log_2\frac{\Omega_s}{\Omega_p},0\right)
\end{equation}
Here $\boldsymbol\Omega_s$ is the solid angle of sampled rays direction, which is
\begin{equation}
    \Omega_s = \frac{1}{Np(\boldsymbol\omega)}
\end{equation}
where $N$ is the number of samples, and $p(\boldsymbol\omega)$ is the pdf. 

\begin{eqnarray}
    \Omega_p = \frac{4}{W^2} \partial A\\
    \partial A = \frac{1}{(u^2+v^2+1)^{3/2}}
\end{eqnarray}
where $u,v\in[-1,1]$ is the projected coordinated of $\boldsymbol\omega$, and $W$ is the resolution of base mipmap level. We add additional lighting smooth loss by minimizing the residual between mipmap level 0 and 1. 
% \subsection{Loss Function}
% We employ L1 loss for rendering loss 
% In NeRFactor, render loss is MSELoss, while smooth loss is L1loss. We however use the contrary. We use L1 loss for the render, while L2 loss for smooth. Our observation is that as we did'nt model inter-reflection as so on, the fitted field cannot fully explain the image. 

\section{Additional Evaluation}
\subsection{Surface Extraction from NeRF}
\begin{figure*}[ht]
    \centering
    \includegraphics[width=0.8\textwidth]{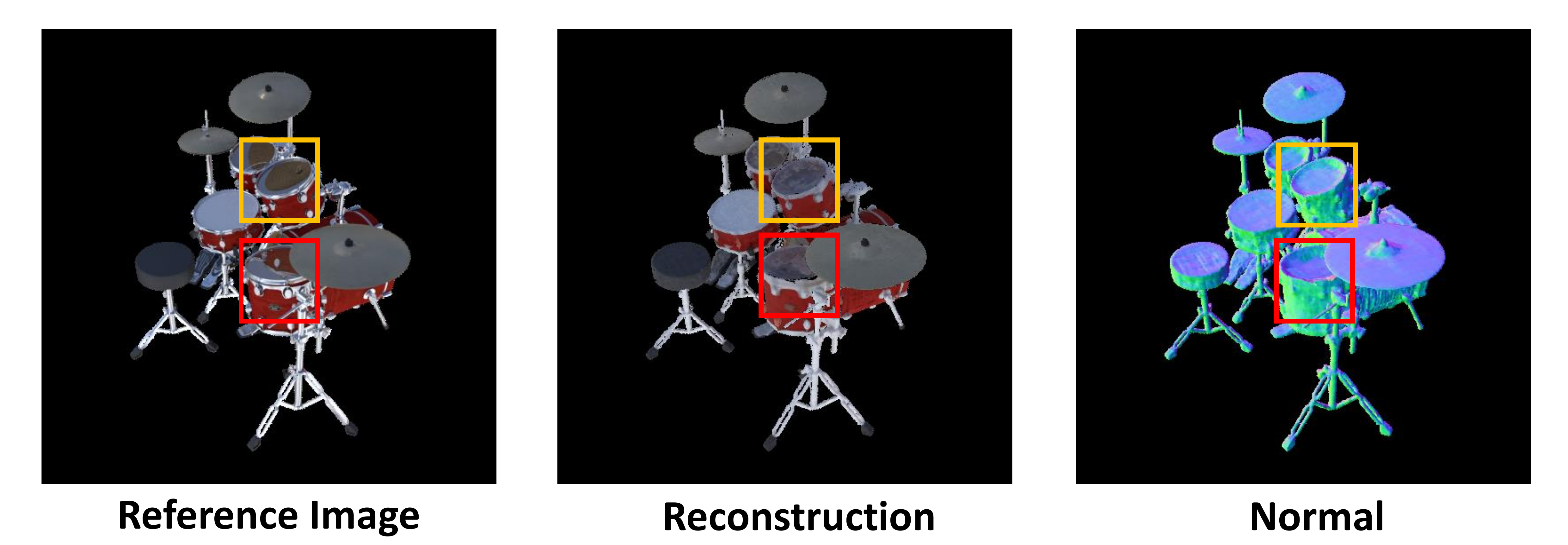}
    \caption{A failure case. Red box: the initial surface from NeRF incorrectly put the bottom part of the drum on the top, and NReF failed to fix this with normal variation only. Yellow box: the top part of the drum is a glass-like transparent surface which violates our opaque surface assumption. Note the original data provided in~\cite{zhang2021nerfactor} masked out the bottom part of the drum in the red box and produces a hollow albedo in that region; we keep this part of mask in this figure to show a typical failure case only.}
    \label{fig:failure}
\end{figure*}
The initial surface normal induced by the surface extraction method in~\cite{mildenhall2020nerf}\cite{zhang2021nerfactor} (Equ.6 in the main paper) is often noisy and erratic.
The reason is that the original shape inducted from NeRF inevitably creates small, double-layered translucent surfaces with vacuumed density in-between, which violates the assumption of opaque surfaces. 
This phenomenon itself, and how it affects the normal induction, are illustrated in figure~\ref{fig:method_visualize_T}.
In theory, an opaque surface should have a step-wise transmittance function along the ray direction (red curve $T_{ideal}(t)$ in fig.~\ref{fig:method_visualize_T}) where the step-change point corresponds to the surface point.
In practice, due to the intrinsic entanglement of surface geometry and view-dependent color effects, the transmittance function trained by NeRF usually comes up with two smaller "step-wise" transitions close to each other (blue curve in fig.~\ref{fig:method_visualize_T}) to compensate for view-dependent color effects.
The two smaller "step-wise" transitions of NeRF corresponds to a double-layered translucent surface locally. Applying Equ.6 in such a case will scatter bad approximation of surface points at the middle vacuumed region. 
These scattered points not only produces noisy surface points itself, but also provides noisy and erratic normal because points in this region usually have a small, unstable gradient towards zero.
Our idea to address this issue is simple - we just make sure the approximated surface point stops at one of the two step-changed points by raymarching with Equ.7 in the main paper. In another word, we find the surface point by performing ray-marching until the transmittance equals to $0.5$.
Since both step-changed points have a well-defined gradient, we can extract a more reasonable surface normal.

\subsection{Parsimony Prior}
Our parsimony term enables long-range/global color similarity. This is demonstrated in Fig.~\ref{fig:parsi}. 
\begin{figure*}[!ht]
    \centering
    \includegraphics[width=0.8\linewidth]{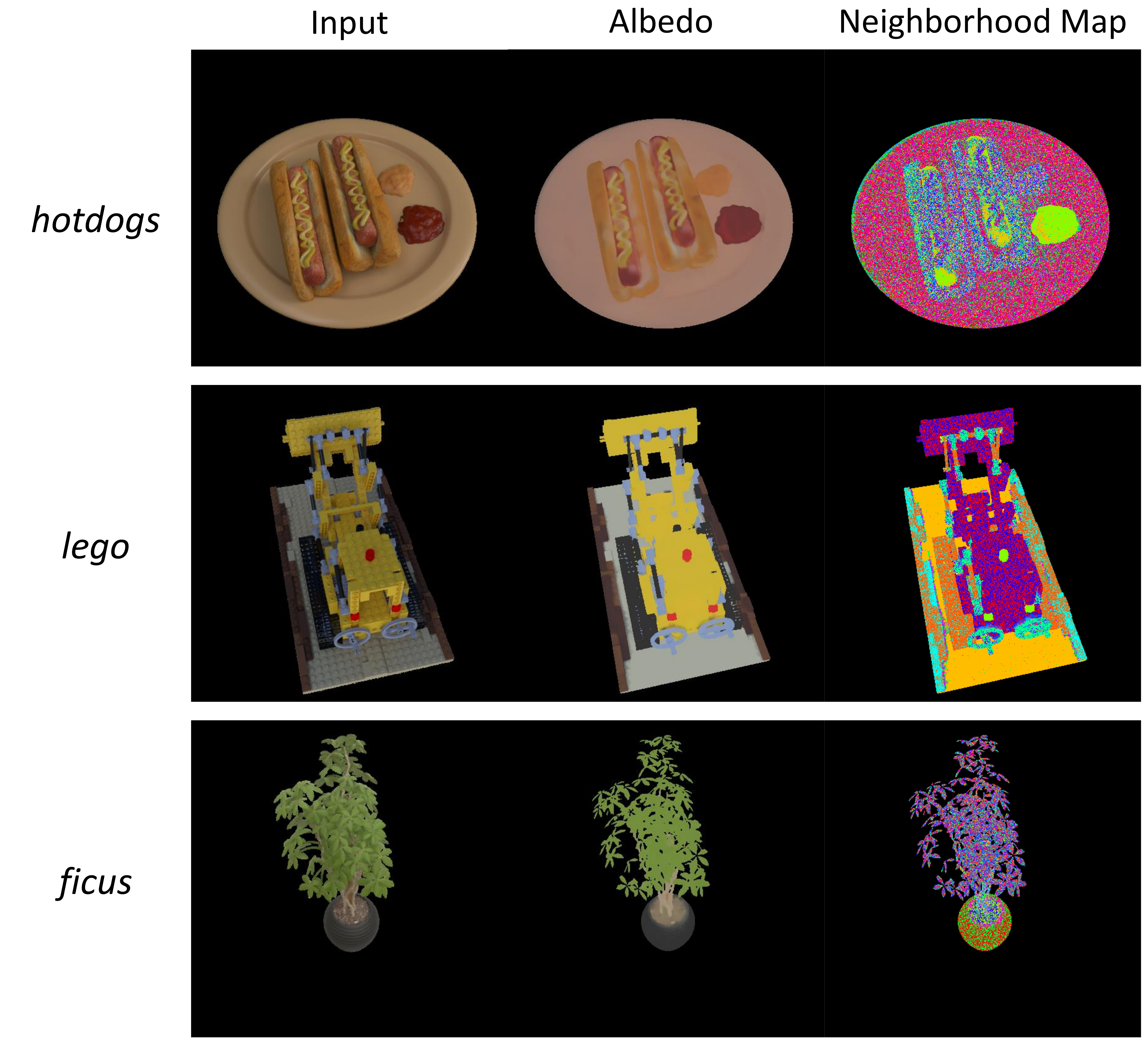}
    \caption{Visualization of the parsimony prior. Left column: input image for reference. Middle column: decomposed albedo map. Right column: visualization of sampled sparse neighbors for each point. The cluster of sampled neighbor is encoded as hsv colors. The noise is due to stochastic sampling.}
    \label{fig:parsi}
\end{figure*}

\section{Additional Results}
\begin{figure*}[!ht]
    \centering
    \includegraphics[width=0.95\linewidth]{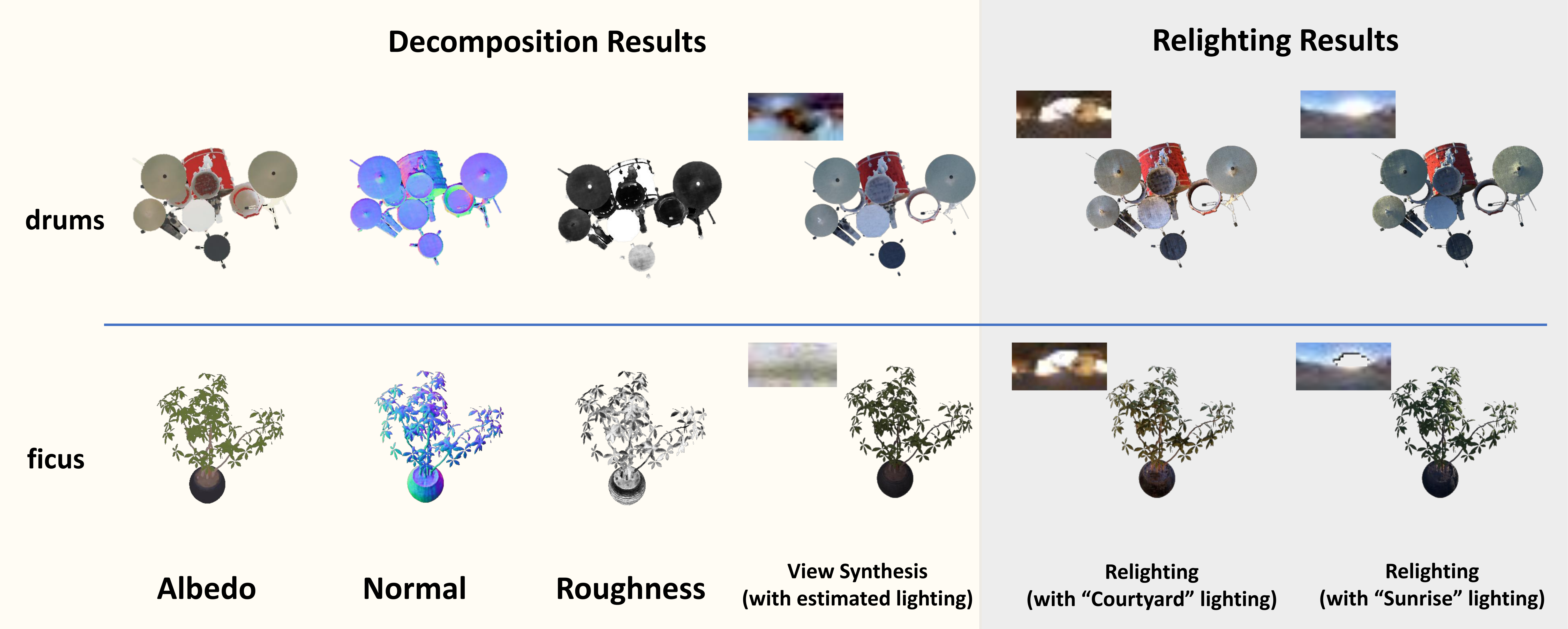}
    \caption{Our full decomposition, reconstruction, and novel relighting results.}
    \label{fig:results_supp}
\end{figure*}
\begin{figure*}[!ht]
    \centering
    \includegraphics[width=0.9\linewidth]{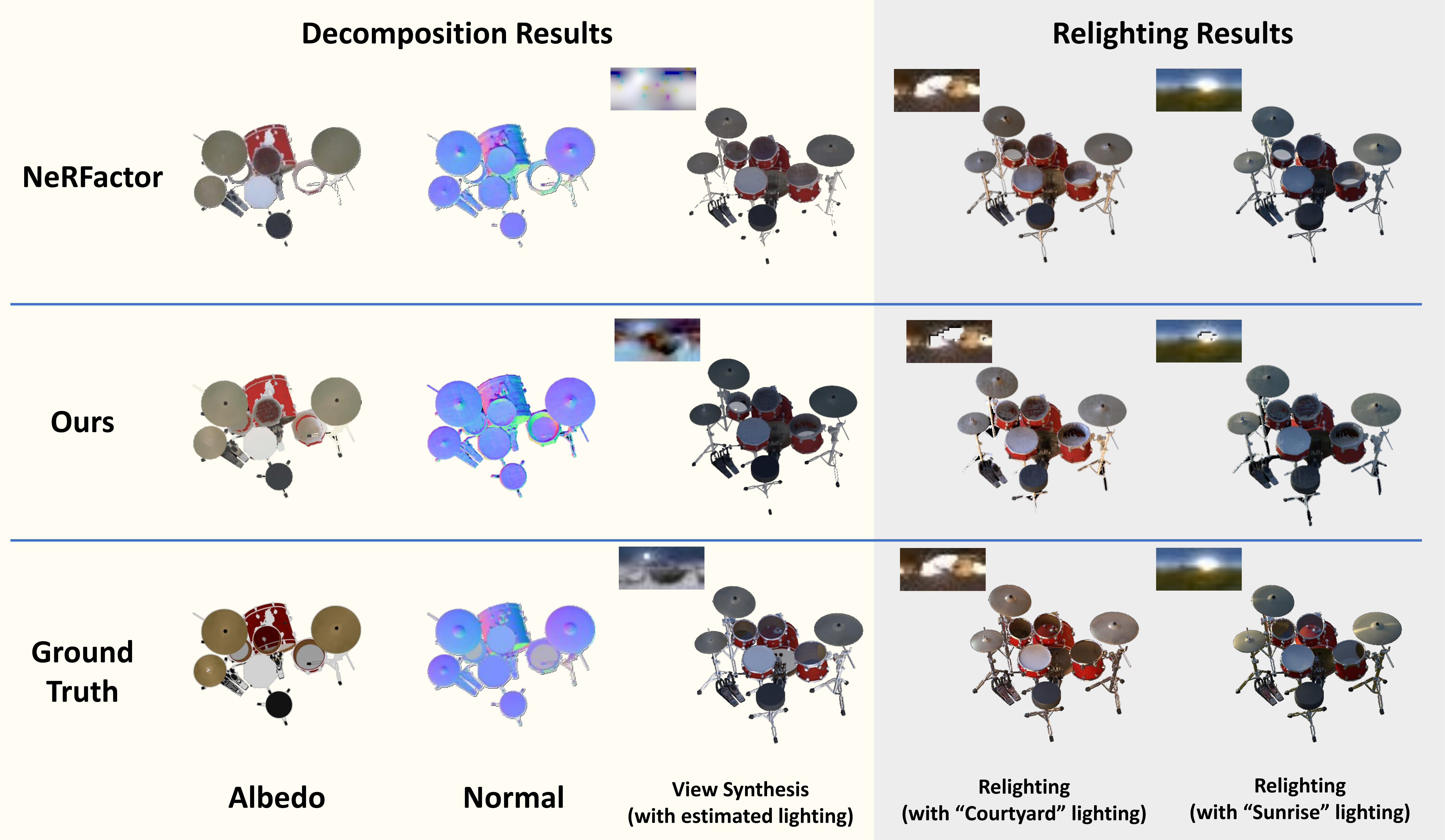}
    \caption{Comparsion of our method with NeRFactor and ground truth on \textit{drums} synthetic data.}
    \label{fig:comparsion_drums}
\end{figure*}

\begin{figure*}[!ht]
    \centering
    \includegraphics[width=0.9\linewidth]{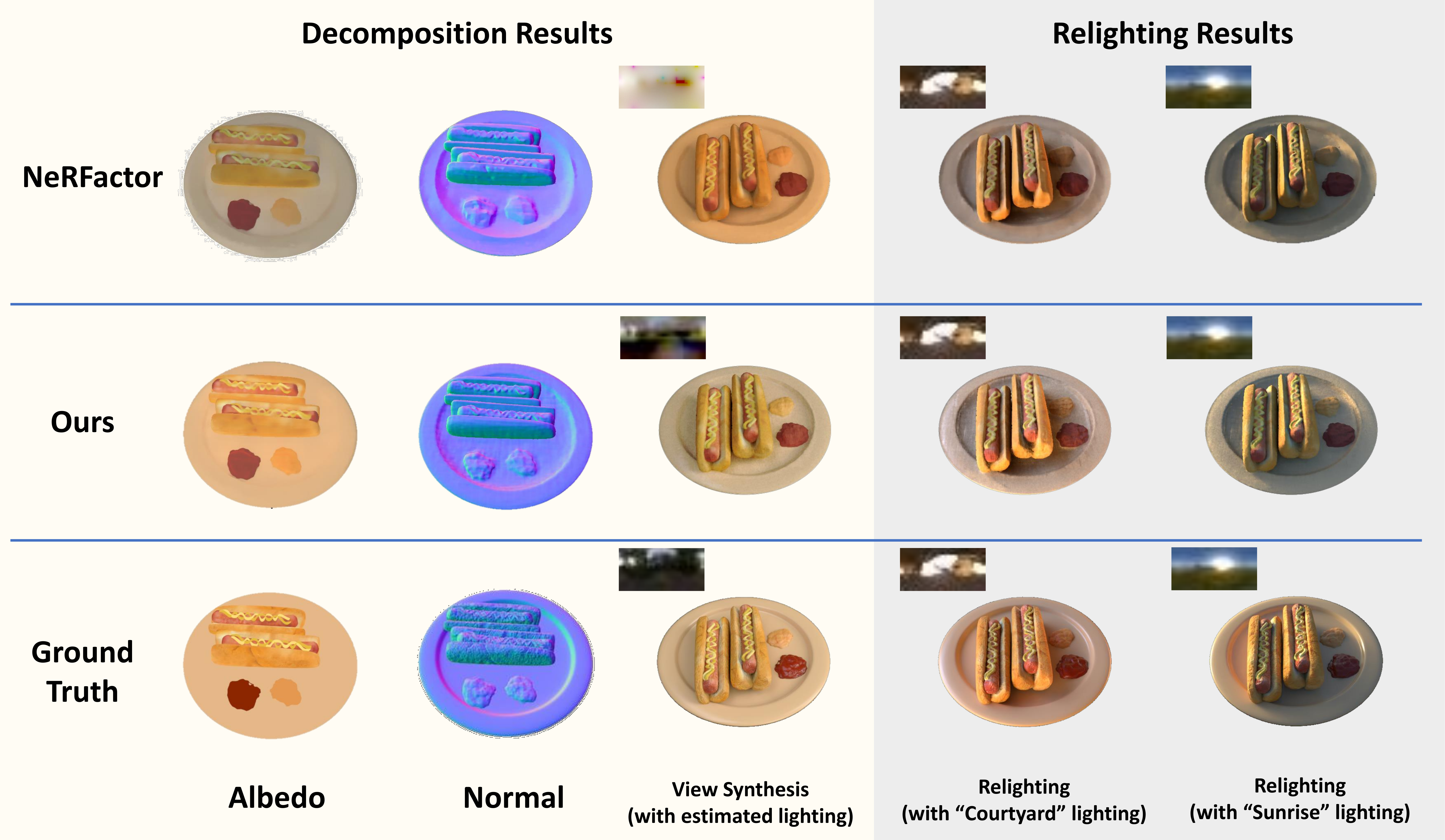}
    \caption{Comparsion of our method with NeRFactor and ground truth on \textit{hotdog} synthetic data.}
    \label{fig:comparsion_hotdog}
\end{figure*}

\begin{figure*}[!ht]
    \centering
    \includegraphics[width=0.9\linewidth]{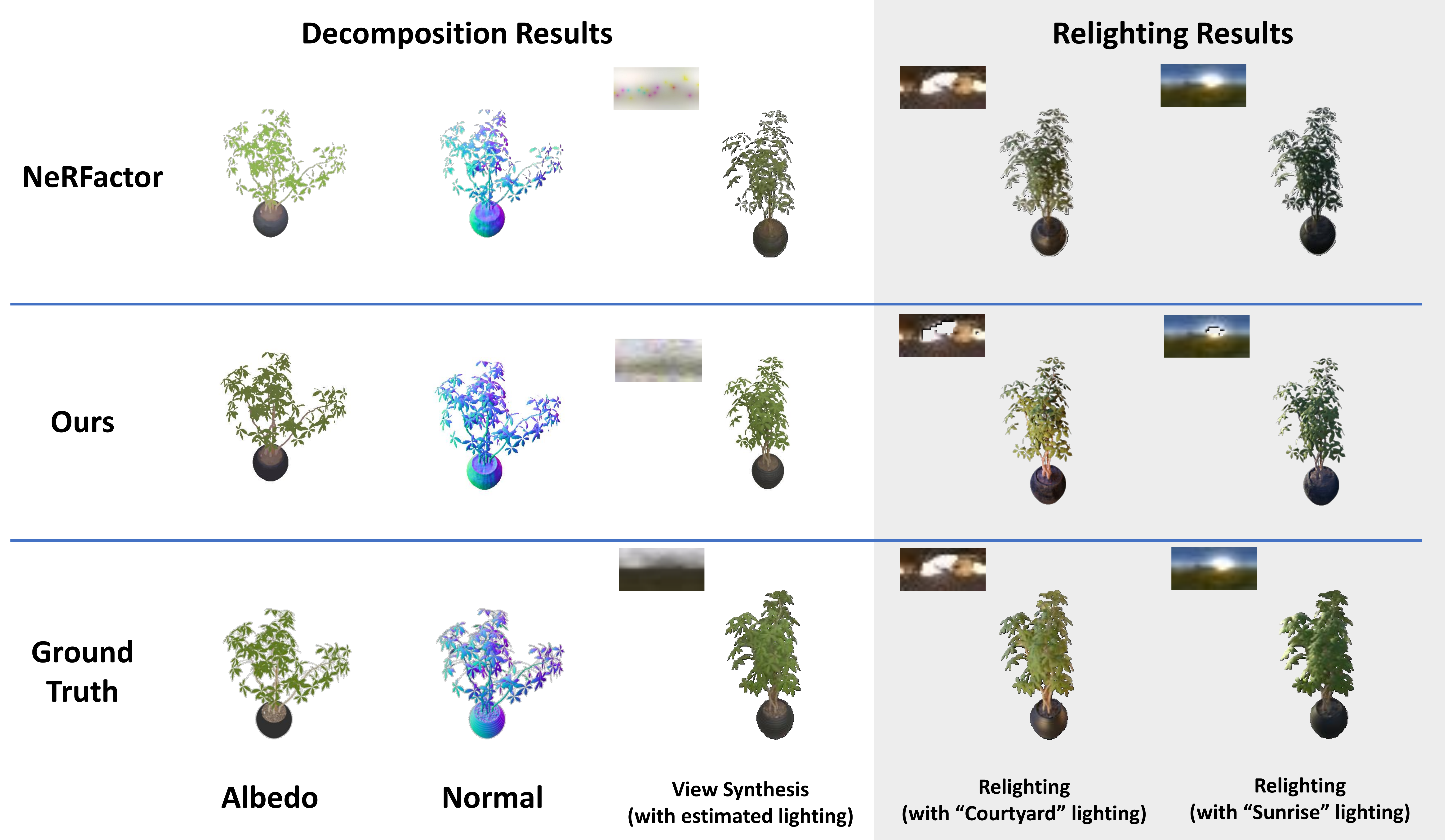}
    \caption{Comparsion of our method with NeRFactor and ground truth on \textit{ficus} synthetic data.}
    \label{fig:comparsion_ficus}
\end{figure*}
\begin{figure*}[!ht]
    \centering
    \includegraphics[width=0.9\linewidth]{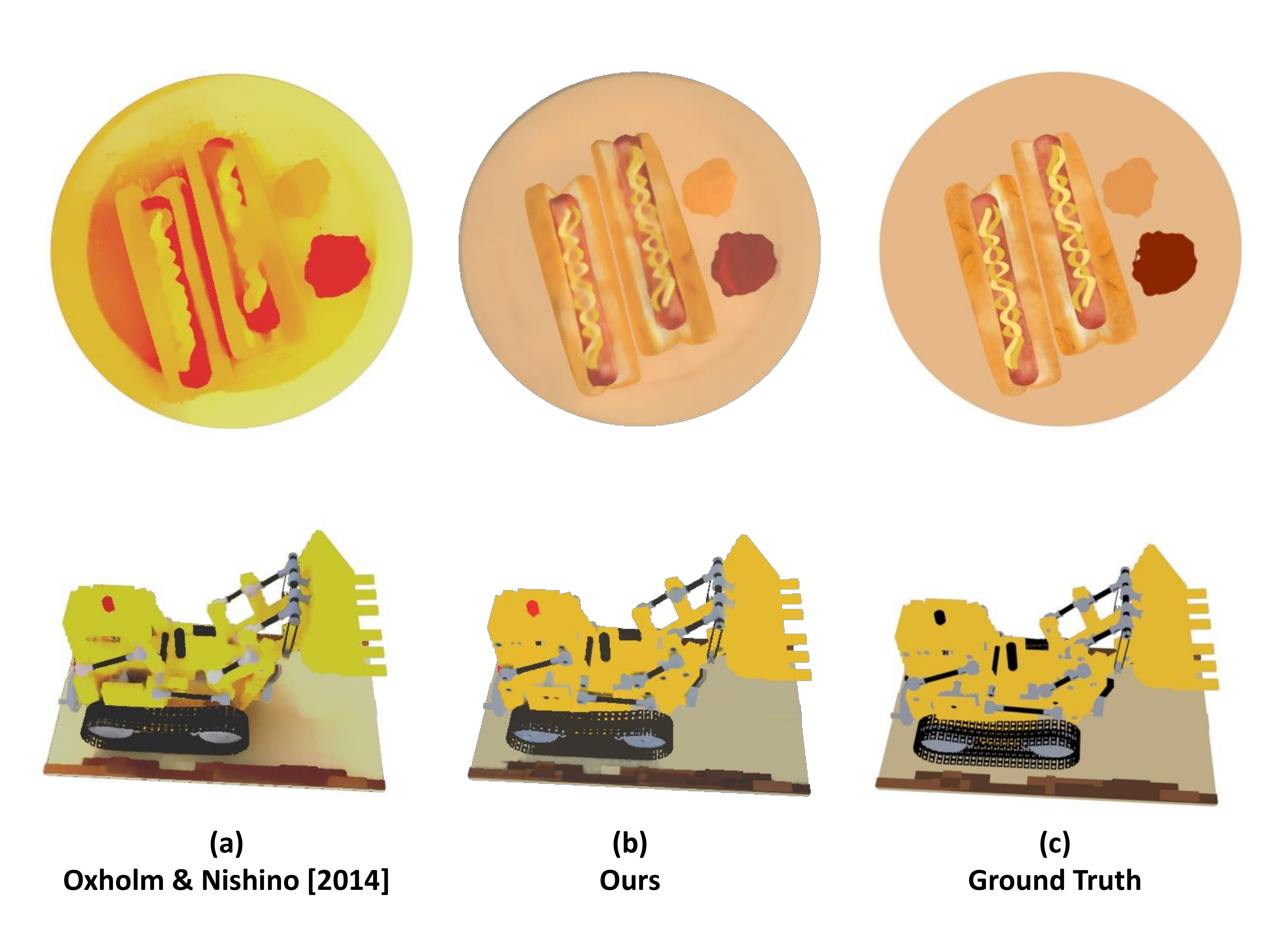}
    \caption{Qualitative Comparison with Oxholm and Nishino~\cite{oxholm2014multiview}. For a fair comparsion, we compare our method with the multi-view enhanced version of~\cite{oxholm2014multiview} from NeRFactor~\cite{zhang2021nerfactor}.
    Our method produce significantly better albedo estimation results with less artifacts.}
    \label{fig:comparsion_nishino}
\end{figure*}
\begin{figure*}[!ht]
    \centering
    \includegraphics[width=\linewidth]{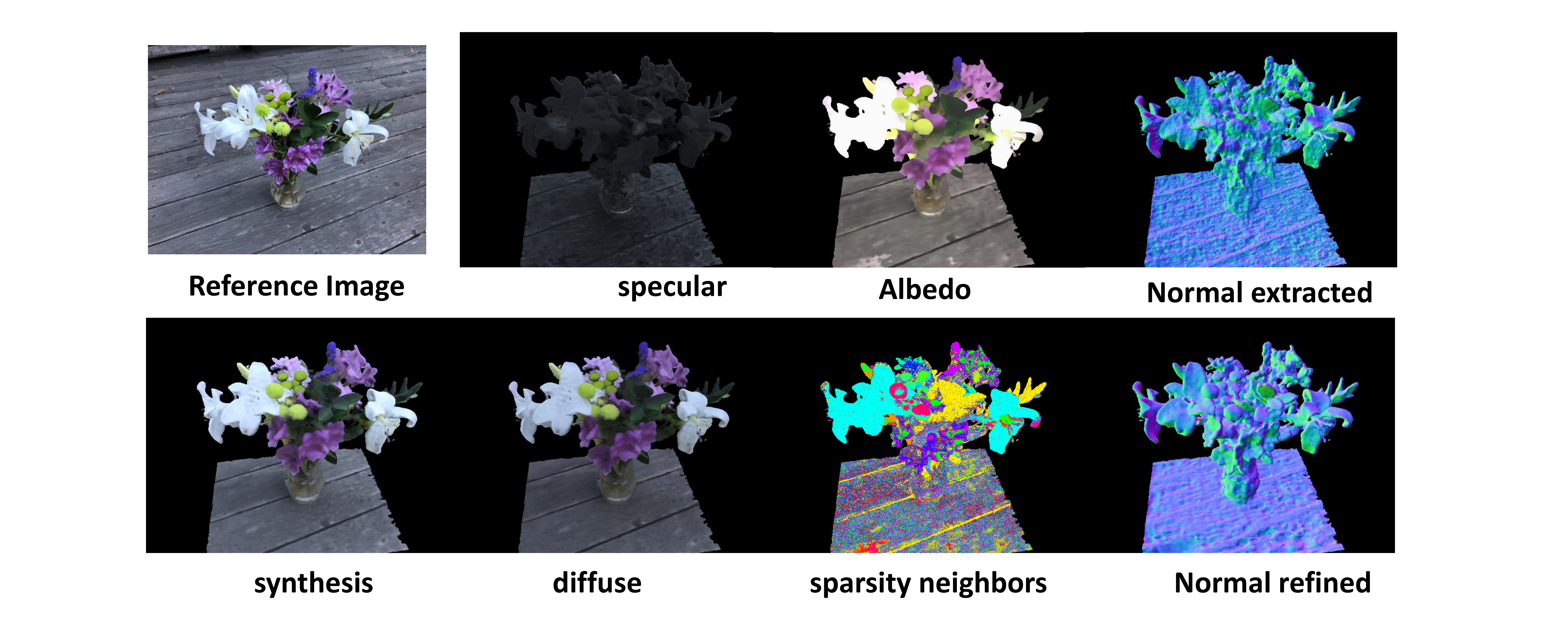}
    \caption{Our full decomposition, reconstruction results on another real captured data.}
    \label{fig:results_supp_real}
\end{figure*}
Here we show additional results on other synthetic data released by~\cite{mildenhall2020nerf} (Fig.~\ref{fig:results_supp}) that are not shown in the main paper, as well as more side-by-side comparison (Fig.~\ref{fig:comparsion_drums}, ~\ref{fig:comparsion_hotdog}, and~\ref{fig:comparsion_ficus}) with NeRFactor~\cite{zhang2021nerfactor} and ground truth.
We also compare our method with previous method of multi-view photometric stereo~\cite{oxholm2014multiview} in fig.~\ref{fig:comparsion_nishino}.
Additional results on real-captured data is shown in fig.~\ref{fig:results_supp_real}.
%Finally, we show that our decomposition method is not limited to use NeRF for initial radiance field estimation: fig. XX shown our decomposition results using other initial reflection field methods than NeRF.

\subsection{Failure cases.}
Our method have a set of assumptions for input, including opaque objects, isotropic materials, inter-reflections free, etc. Violating those assumptions might produces incorrect results.
The NReF relies on initial surfaces extracted from NReF and only refines normal but not surface positions; thus a strongly deviated shape might not be able to be fixed and produces artifacts. 
A typical failure case is shown in Fig~\ref{fig:failure}.

\end{document}